\documentclass[10pt,twocolumn,letterpaper]{article}

\usepackage{iccv}              

\usepackage{algorithm}
\usepackage{algpseudocode}
\usepackage{amsmath}
\usepackage{booktabs}
\usepackage{graphicx}
\usepackage{xcolor}
\usepackage{multirow}
\usepackage{colortbl}
\usepackage{tikz}
\usepackage{multirow}

\definecolor{gold}{HTML}{FBF2D2}
\definecolor{silver}{HTML}{DDDDDD}
\definecolor{bronze}{HTML}{EED2B8}
\definecolor{new_blue}{HTML}{3B82F6}

\definecolor{goldD}{HTML}{D9AE13}
\definecolor{silverD}{HTML}{909090}
\definecolor{bronzeD}{HTML}{9A5F26}

\definecolor{catGreen}{HTML}{238763}
\definecolor{catBlue}{HTML}{1F70AE}

\newcommand{\medal}[3]{\tikz[baseline=(char.base)]{\node[rounded corners=2pt,fill=#1,draw=#2,inner sep=1.5pt] (char) {#3};}}

\newcommand{\bm}[2]{
    \ifcase#1\or
      {\medal{gold}{goldD}{\textbf{#2}}}
    \or 
      {\medal{silver}{silverD}{#2}}
    \or 
      {\medal{bronze}{bronzeD}{#2}}
    \else 
      #2
    \fi\ignorespaces
}

\definecolor{iccvblue}{rgb}{0.21,0.49,0.74}
\usepackage[pagebackref,breaklinks,colorlinks,allcolors=iccvblue]{hyperref}

\def\paperID{****}
\def\confName{CVPR}
\def\confYear{2026}

\title{AnomalyVFM -- Transforming Vision Foundation Models into Zero-Shot Anomaly Detectors}

\author{Matic Fučka$^{1,}$\thanks{Equal contribution.} \and Vitjan Zavrtanik$^{1,2,}$\footnotemark[2] \and Danijel Skočaj$^{1}$ \and
$^{1}$University of Ljubljana, Faculty of Computer and Information Science, Slovenia \\
$^{2}$*codeplain
\\
{\tt\small \{matic.fucka, vitjan.zavrtanik, danijel.skocaj\}@fri.uni-lj.si}
}

\begin{document}
\maketitle
\begin{abstract}
Zero-shot anomaly detection aims to detect and localise abnormal regions in the image without access to any in-domain training images. While recent approaches leverage vision–language models (VLMs), such as CLIP, to transfer high-level concept knowledge, methods based on purely vision foundation models (VFMs), like DINOv2, have lagged behind in performance. We argue that this gap stems from two practical issues: (i) limited diversity in existing auxiliary anomaly detection datasets and (ii) overly shallow VFM adaptation strategies. To address both challenges, we propose AnomalyVFM, a general and effective framework that turns any pretrained VFM into a strong zero-shot anomaly detector. Our approach combines a robust three-stage synthetic dataset generation scheme with a parameter-efficient adaptation mechanism, utilising low-rank feature adapters and a confidence-weighted pixel loss. Together, these components enable modern VFMs to substantially outperform current state-of-the-art methods. More specifically, with RADIO as a backbone, AnomalyVFM achieves an average image-level AUROC of $94.1\%$ across 9 diverse datasets, surpassing previous methods by significant $3.3$ percentage points. 
\href{https://maticfuc.github.io/anomaly_vfm/}{Project Page}
\end{abstract}    
\section{Introduction}
\label{sec:intro}

\begin{figure}[htb]
    \centering
    \includegraphics[width=1.0\columnwidth]{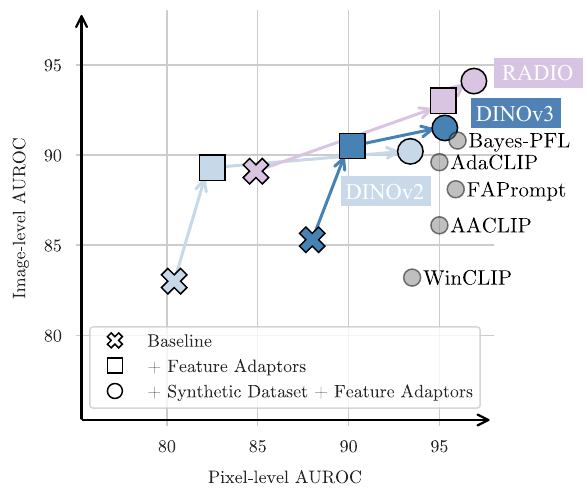}
    \caption{Vision–language models excel in zero-shot anomaly detection thanks to their high-level concept knowledge, but purely visual foundation models hold untapped potential. AnomalyVFM unlocks this potential by addressing the two practical limitations that hinder VFM underperformance: suboptimal training sets and suboptimal fine-tuning procedures.}
    \label{fig:main}
    \vspace{-0.6cm}
\end{figure}

Visual anomaly detection aims to identify abnormal regions at test time while training only on anomaly-free images. This represents a foundational task within manufacturing~\cite{bergmann2019mvtec, visa2022, wang2024real}, medical imaging~\cite{salehi2021multiresolution,jha2019kvasir,hamada2020br35h} and road obstacle detection~\cite{vojivr2023image,vojivr2024pixood,Delic_2024_BMVC}. In industrial inspection, it is typically assumed~\cite {fuvcka2025transfusion, roth2022towards} that many normal images are available during training. However, practical deployments often require detecting anomalies on arbitrary object classes without any or very few images. This extremely challenging setting has motivated recent interest in few-shot and zero-shot anomaly detection. Few-shot methods~\cite{lipromptad_fewshot,zhu2024toward,tao2025kernel} require a handful of normal images of the object class, while zero-shot methods~\cite{BayesFPL, cao2024adaclip, zhou2024anomalyclip} must generalise to unseen object classes with no in-domain images at all.

State-of-the-art zero-shot approaches~\cite{zhou2024anomalyclip, cao2024adaclip, BayesFPL} use vision–language models (VLMs) such as CLIP~\cite{radford2021learning}. They typically use auxiliary anomaly detection datasets to train the model to output text embeddings that encode generic notions of normality and abnormality. Pretraining with image-text supervision introduces valuable high-level concept knowledge, which facilitates generalisation across different object categories. By contrast, pure vision foundation models (VFMs) such as DINOv2~\cite{oquab2023dinov2} encode strong visual representations but have so far trailed behind VLM-based methods when used as a basis for zero-shot anomaly detection. This gap raises the question: Can VFMs, which are arguably better suited to the fundamentally visual nature of anomaly detection, be transformed into competitive zero-shot detectors?

We argue that two practical limitations explain why VFMs have underperformed in prior zero-shot work. First, existing auxiliary anomaly datasets~\cite{bergmann2019mvtec,visa2022} lack sufficient diversity and coverage of realistic defects, which are required for training VFMs. This is not a problem for VLMs due to their high-level concept knowledge. When the model must generalise to arbitrary object classes, limited dataset diversity prevents learning broadly applicable cues. Second, most prior VFM adaptations~\cite{cao2024adaclip,cao2023segment,oquab2023dinov2} fine-tune only a small output head with simple pixel-wise losses, leaving the model's internal visual representations essentially unchanged. This makes it challenging for the model to accurately learn the features necessary to distinguish between normal and abnormal appearances across various objects.

To address both points, we propose AnomalyVFM, a practical framework that transforms any modern VFM into a robust zero-shot anomaly detector. AnomalyVFM has two core components. First, a three-stage synthetic dataset generator that uses modern image generation models (e.g., FLUX~\cite{flux2024}) to (i) create diverse anomaly-free object images, (ii) synthesise a wide variety of local defects by inpainting at sampled locations, and (iii) filter generated samples using a feature-based verification step to ensure the presence and relevance of defects. This creates a large and diverse auxiliary training set containing many object/background combinations. Second, we introduce a parameter-efficient adaptation mechanism tailored for VFMs: low-rank feature adapters are injected throughout the (transformer) backbone, coupled with a lightweight decoder and a confidence-weighted pixel loss that downweights ambiguous supervision. The adapters enable the VFM to evolve its internal visual representations (not just the final head) with minimal additional parameters. The decoder converts these adapted features into pixel-level anomaly scores, and the confidence-weighted loss limits the impact of noisy gradients from imperfect synthetic labels. Crucially, AnomalyVFM is model-agnostic and practical: it can be applied to any pretrained VFM with a transformer backbone (as shown in Figure~\ref{fig:main}).

In summary, our contributions are:
\begin{itemize}
\item As our main contribution, we introduce AnomalyVFM, an effective framework that transforms any pretrained VFM into a competitive zero-shot anomaly detector using a parameter-efficient adaptation scheme and a synthetically generated dataset containing diverse data.
\item As our secondary contribution, we propose a scalable scheme for generating synthetic anomaly detection datasets. We design a three-stage synthesis process that leverages modern generative models to produce diverse object instances, realistic local defects, and automatic feature-based verification to ensure data quality. This yields data that is better suited for finetuning VFMs in comparison to existing datasets.
\end{itemize}
We validate our contributions by evaluating the proposed approach on nine standard industrial anomaly detection benchmarks, surpassing the previous best zero-shot anomaly detection methods by a significant 3.3 percentage points (p.\ p.) in image-level AUROC and 0.9 p.\ p.\ in pixel-level AUROC. Additionally, we demonstrate that AnomalyVFM also generalises well on medical anomaly detection benchmarks, even though it was not finetuned for this purpose. Finally, we demonstrate the versatility of AnomalyVFM by finetuning it on a few normal samples. With this, it is able to match the performance of state-of-the-art models in the few-shot regime.

\section{Related Work}
\label{sec:related}

\noindent \textbf{Anomaly detection} can be categorized into several paradigms. Most commonly they are divided in reconstructive~\cite{zavrtanik2021reconstruction,ae-ssim}, discriminative~\cite{zavrtanik2021draem,zavrtanik2022dsr,liu2023simplenet,fuvcka2025transfusion} and embedding-based methods~\cite{roth2022towards, defard2021padim}. Reconstructive approaches~\cite{batzner2024efficientad, pirnay2022inpainting} are trained to reconstruct anomaly-free images. Since the learnt models never see anomalous examples, it is assumed that they will be poorly reconstructed, making them detectable via reconstruction error. Discriminative methods~\cite{supersimplenetv2,salad} are trained with synthetic anomalies under the assumption that this will generalise well to actual anomalies. Embedding-based methods~\cite{deng2022anomaly, zhou2024msflow,fuvcka2026objectcore} fit a simple normality model, such as a coreset, on top of the features extracted from a pretrained encoder.

\noindent\textbf{Training from generated data} is common in solving or evaluating text-based tasks~\cite{borisov2022languagesyn,whitehouse2023llmpoweredsyn,yoo-etal-2021-gpt3mix-leveragingsyn,preda2024supporting}. It has also started being adapted in computer vision in areas such as video generation~\cite{Zhao_2025_ICCV} and dynamic scene reconstruction~\cite{Chen_2025_ICCV}, but has not yet seen widespread usage. For training data generation, powerful diffusion models~\cite{rombach2022high} or flow-matching~\cite{esser2024scaling} approaches are commonly used~\cite{karazija2024diffusion,liu2024can,kupyn2024dataset}. In \cite{liu2024can}, a diffusion model is utilised to generate samples according to a set of labels for out-of-distribution object detection. The method focuses on outlier generation across different object classes but overlooks near-in-distribution cases. 

\noindent\textbf{Anomaly synthesis} has started to receive more attention in the past few years. The field was started by DR{\AE}M~\cite{zavrtanik2021draem}, which synthesised anomalies by cropping and pasting parts of images from an external dataset~\cite{dtdsynth}. Some approaches later improved upon this by refining how external images are augmented or by moving synthetic anomaly generation to the latent space~\cite{liu2023simplenet, supersimplenet, zavrtanik2024cheating}. Some of the later approaches improved upon the realism of the generated samples by using modern generative models, such as generative adversarial networks~\cite{Zhang_DefectGAN,Duan_gan} of diffusion models~\cite{yang_defect_spectrum,Zhang_AnomalyDiffusion}. However, all of these methods typically require substantial amounts of normal and/or abnormal data. Furthermore, these methods can only generate samples similar to the training set, i.e., seen anomalies, but fail to generate unseen anomalies. This makes them unsuitable for zero-shot anomaly detection. Unlike previous approaches, our generation approach does not require any samples, normal or abnormal.

\noindent \textbf{Zero-shot anomaly detection} methods detect anomalies at inference while never seeing an instance of the observed object during training. Most recent zero-shot anomaly detection methods~\cite{jeong2023winclip,zhou2024anomalyclip, zhu2024llms, chen2024clip, cao2024adaclip} focus on utilising the general object appearance knowledge embedded in vision-language~\cite{radford2021learning} models. 
A minority of methods have utilised Vision (only) Foundation Models for the task of zero-shot anomaly detection. SAA~\cite{cao2023segment} uses the GroundingDINO~\cite{liu2024grounding} and SAM~\cite{kirillov2023segment} with handcrafted anomaly prompts to directly segment anomalies. In \cite{li2024zero}, the method models the distribution of object appearance within the batch to detect anomalous samples. However, assumptions about the contents of a specific batch are often violated in real-world scenarios. Some methods~\cite{cao2024adaclip, oquab2023dinov2} have also investigated tuning VFMs directly on an auxiliary dataset, but achieved suboptimal results.

In this paper, we demonstrate that it is possible to achieve state-of-the-art performance using a pretrained VFM finetuned on a sufficiently diverse dataset.

\section{Dataset Generation Scheme} \label{sec:data_gen}

\begin{figure*}
    \centering
    \includegraphics[width=\linewidth]{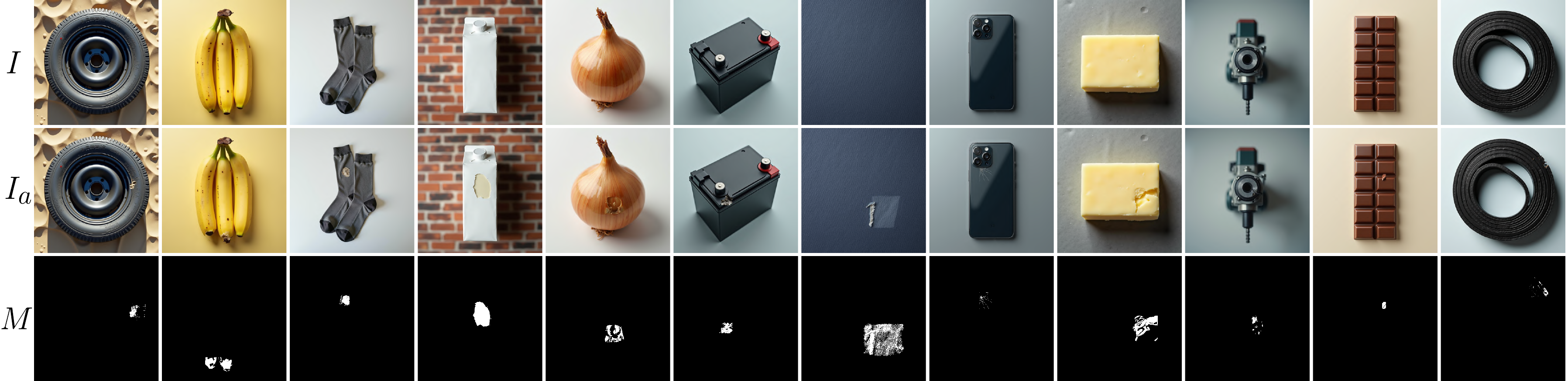}
    \caption{Examples of generated anomaly-free images $I$, anomalous images $I_a$ and corresponding masks $M$.
    }
    \label{fig:defects}
    \vspace{-0.2cm}    
\end{figure*}

To address the issue of data diversity, a collection of realistic images of objects, both with and without anomalies, is necessary. Additionally, each anomalous image should be accompanied by a pixel-level annotation. To do this, a three-stage generation scheme is proposed. (i) First, the initial image of the object is generated. (ii) Then, a realistic defect is inpainted on top of the object. (iii) Ultimately, the anomaly segmentation map is generated by subtracting the features of the normal image from those of the anomalous image, and based on this, poorly generated images are filtered. Each of these steps will be described in detail below. Some examples of generated samples are shown in Figure~\ref{fig:defects}.

\begin{figure}
    \centering
    \includegraphics[width=\columnwidth]{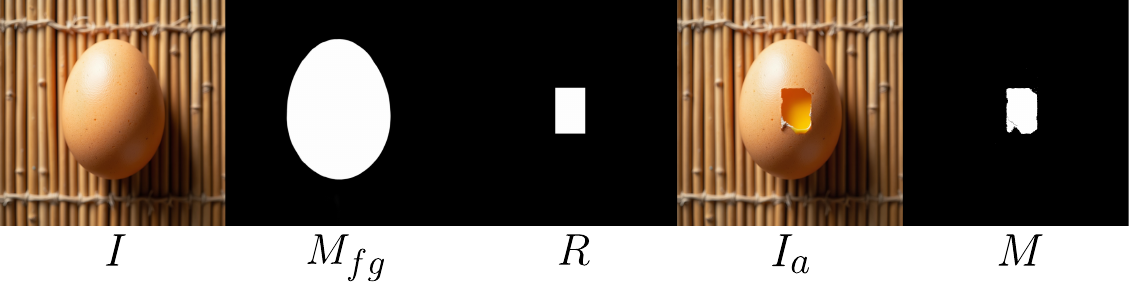}
    \caption{Dataset generation pipeline. The image $I$ is generated using a text-conditioned image generation model. Then, the foreground mask $M_{fg}$ is extracted and an anomalous region $R$ is sampled from it. Then, the anomalous image $I_a$ is generated by inpainting an anomaly inside $R$. Finally, features are extracted from $I$ and $I_a$, and then compared and thresholded to obtain $M$.
    }
    \label{fig:anogen}
    \vspace{-0.3cm}
\end{figure}

\noindent \textbf{Anomaly-free Image Generation} To generate the initial (anomaly-free) image $I$ (Figure~\ref{fig:anogen}, $I$), an image generation model $G$ is prompted with an anomaly-free text prompt $p$:
\begin{equation}
    I = G(p).
\end{equation}
As the image generation model $G$, the flow-matching-based FLUX model~\cite{flux2024} is used in all experiments unless stated otherwise. Anomaly-free text prompt $p$ is constructed as follows: \\

\noindent \texttt{A close-up photo of [Object] for industrial visual inspection. Top-down view. Centered. [Texture] background.} \\

\noindent The \texttt{[Object]} and \texttt{[Texture]} are replaced by an object or background class from a list of 100 objects and 50 backgrounds generated by an LLM (in our case GPT-4o~\cite{gpt4o}).

\noindent \textbf{Anomalous Image Generation} The anomalous image $I_a$ is generated using the anomaly-free image $I$. To do so, the rough anomaly location $R$ (in our case, a rectangle) has to be determined. As the first step, the foreground object mask $M_{fg}$ must be extracted. In our case, this is done using a pretrained salient object segmentation network IS-Net~\cite{qin2022highly} (Figure~\ref{fig:anogen}, $M_{fg}$). To generate $R$, the location of the anomaly $(x,y)$ is first sampled on the foreground object, i.e., as a random positive pixel in $M_{fg}$. The initial location serves as the centre of the anomaly rectangle $R$ (Figure~\ref{fig:anogen}, $R$). The width and height are uniformly sampled according to the desired anomaly width $(w_{min}, w_{max})$ and height $(h_{min}, h_{max})$ parameters:
\begin{equation}
    w \sim \text{U}(w_{min}, w_{max}), h \sim \text{U}(h_{min}, h_{max}).
\end{equation}

The anomaly is then generated by prompting the model $G$ with an anomalous prompt $p_a$, while restricting the generation to the region $R$ and maintaining $I$ in other regions, i.e. inpainting. To generate an anomalous version of the generated image $I$, the prompt $p_a$ additionally contains anomalous descriptions: \\

\noindent \texttt{A close-up photo of a [Anomaly] [Object] for industrial visual inspection. Top-down view. Centered. [Texture] background.} \\

\noindent The \texttt{[Anomaly]} tag is replaced with a description of an anomaly, such as \texttt{cracked, damaged, smudged, rotten}. A list of \texttt{[Anomaly]} descriptions for each \texttt{[Object]} is generated by an LLM (again GPT-4o~\cite{gpt4o}). This ensures the \texttt{[Anomaly]} is relevant for the object. The \texttt{[Object]}, \texttt{[Anomaly]} and \texttt{[Texture]} lists are listed in the Supplementary material.

No inpainting-specific models are used; instead, the characteristics of the iterative generation process (diffusion or flow matching) are used, using the RePaint approach~\cite{lugmayr2022repaint}. 
The iteratively generated image $I_a$ (Figure~\ref{fig:anogen}, $I_a$) contains the object generated in $I$ with an anomaly in region $R$ whose visual appearance follows the prompt in $p_a$. However, accurate prompt adherence is not a solved problem in image generation~\cite{esser2024scaling}, so some generated images may not contain anomalies at all. To address this, a filtering process is proposed which removes the vast majority of examples where anomalies are not generated in $I_a$.

\noindent \textbf{Dataset filtering} To filter out examples with poor adherence to $p_a$, a comparison between the anomaly-free $I$ and the corresponding anomalous $I_a$  is performed. First, DINOv2~\cite{oquab2023dinov2} features are extracted from $I$ and $I_a$, obtaining $f$ and $f_a$, respectively. The extracted features are then compared using cosine distance, obtaining a distance map $M_d$. The maximum value of $M_d$ is obtained as the distance score $D$. The mask is binarised according to a threshold $T$ to obtain the final mask $M$. The generated sample is accepted if the distance $D$ exceeds a set threshold $T$. The idea behind this filtering process is that if the anomaly generation step fails to adhere to $p_a$, an anomaly-free object region will be generated in region $R$ instead of the anomaly. In this case, the inpainted region will be closer to the anomaly-free object appearance distribution, so the distance between $f$ and $f_a$ should be smaller, enabling the detection of failed examples.

Generated triplets containing an anomaly-free image $I$, an anomalous example $I_a$, and the corresponding anomaly mask $M$ can be seen in Figure~\ref{fig:defects}.

\section{AnomalyVFM} \label{sec:ano_det}

\begin{figure*}
    \centering
    \includegraphics[width=\textwidth]{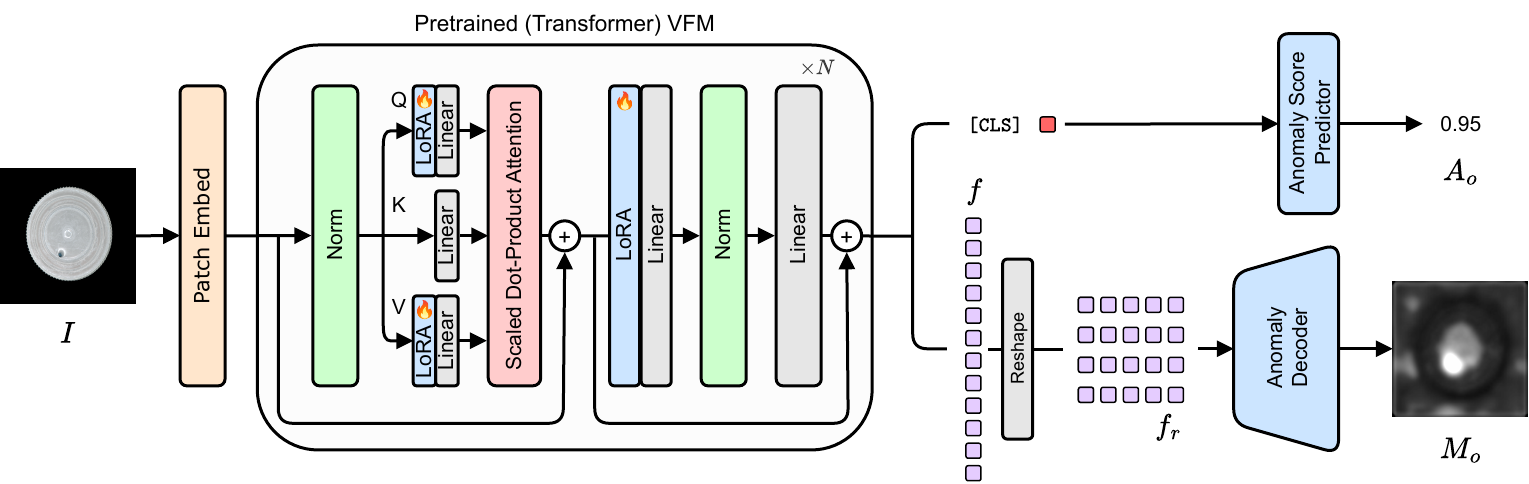}
    \caption{Architecture of AnomalyVFM. All additions to the base VFM are colored in \textcolor{new_blue!80}{blue}.
    }
    \label{fig:architecture}
    \vspace{-0.3cm}
\end{figure*}

Recent attempts~\cite{cao2024adaclip} at using VFMs for zero-shot anomaly detection have only appended a simple MLP on top of the method to generate an anomaly mask, disregarding the adaptation of internal values, the design of the decoder and the losses used to train them. To improve upon this, an effective and parameter-efficient finetuning technique is proposed. First, we improve upon the decoder and inject feature adaptation modules within the VFM to enable adaptation of internal layers. In addition, we propose a confidence-weighted loss to mitigate potential ambiguity caused by inaccurate labels. The adaptation network and the confidence-weighted loss will be described in detail below. The architecture of AnomalyVFM is shown in Figure~\ref{fig:architecture}.

\noindent \textbf{Feature Adaptation Module and Decoder} The input image $I$ is input into the pretrained backbone $F$. Each transformer block $b$ of $F$ is integrated with a Feature Adaptation Module. More specifically, we integrate a LoRA~\cite{hu2022lora} block into the attention mechanism~\cite{vaswani2017attention} by injecting it into the query, value, and output projection layers, as shown in Figure~\ref{fig:architecture}. If not stated otherwise, the rank of LoRA is equal to 64. The extracted features $f$ from the final block of the backbone $F$ are reshaped to $f_r$. Then, $f_r$ is input into a small convolutional decoder that upsamples the features. The decoder is composed of two sequential upsampling blocks, which are constructed as a convolutional layer, a GroupNorm layer~\cite{wu2018group}, a ReLU activation function, and a bilinear upsampling operation. A final Convolutional Layer is used to output both the output anomaly segmentation map $M_o$ and the confidence map $c$. The \texttt{[CLS]} tokens of the backbone are input into a simple linear layer, which predicts an image-level anomaly score $A_{o}$.

\noindent \textbf{Confidence-weighted loss} The Feature Adaptation Modules, Anomaly Decoder and Anomaly Score Predictor are trained jointly. For the image-level loss $\mathcal{L}_{img}$, Focal Loss~\cite{lin2017focal} is used, while the base loss for segmentation $\mathcal{L}_{base}$ is a combination of Focal loss and $\mathcal{L}_1$ loss, following recent anomaly detection methods~\cite{yang2023memseg}:
\begin{equation}
    \mathcal{L}_{base} = \mathcal{L}_{1}(M_{o}, M_{GT}) + \beta * \mathcal{L}_{focal}(M_{o}, M_{GT}),
\label{eq:conf}
\end{equation}
where $\beta$ is equal to 5. Additionally, to better handle the noisy segmentation masks that occur during data generation and any ambiguities in the ground truth masks, we weight the loss with the confidence output from the anomaly decoder, similar to 3D reconstruction methods~\cite{wang2024dust3r,kendall2018multi}. More specifically, the segmentation loss is defined as follows:
\begin{equation}
    \mathcal{L}_{seg} = \mathcal{L}_{base}(M_{o}, M_{GT}) * C - \alpha log(C),
\label{eq:conf}
\end{equation}
where $C$ is defined as $C=1 + exp(c)$, where $c$ is the confidence map predicted by the decoder, and $\alpha$ is equal to $0.1$. The full loss is the sum $\mathcal{L} = \mathcal{L}_{seg} + \mathcal{L}_{img}$.

At inference, the image $I$ is passed through the model, which directly returns both the output anomaly segmentation mask $M_{o}$ and the image-level anomaly score $A_{o}$.

\section{Experiments}
\label{sec:experiments}

\begin{table}
\centering
\caption{Generalisation across different VFMs. Improvement over the baseline is shown in \textcolor{ForestGreen}{green}. SD stands for Synthetic dataset, and FA stands for Feature Adaptors. The average results across 9 industrial datasets are reported.}
\label{tab:contr}
\vspace{-8pt}
\resizebox{1.0\columnwidth}{!}{
\begin{tabular}{lcccccc} 
         \toprule
         \multirow{2}{*}{\textbf{Model}} & \multicolumn{2}{c}{Additions}  & \multicolumn{2}{c}{Image-level} & \multicolumn{2}{c}{Pixel-level}  \\
          \cmidrule(r){2-3} \cmidrule(r){4-5} \cmidrule(r){6-7} & SD & FA   & AUROC & $F_1$-Max & AUROC & $F_1$-Max \\
        \midrule
        \multirow{4}{*}{DINOv2~\cite{oquab2023dinov2}} & & & 83.0 & 78.9 & 80.4 & 23.9  \\
        & \checkmark & & 86.4 \textcolor{ForestGreen}{\footnotesize $\uparrow 3.4$} & 79.3 \textcolor{ForestGreen}{\footnotesize $\uparrow 0.4$} & 93.2 \textcolor{ForestGreen}{\footnotesize $\uparrow 12.8$} & 41.2 \textcolor{ForestGreen}{\footnotesize $\uparrow 17.3$} \\
        & & \checkmark & 89.3 \textcolor{ForestGreen}{\footnotesize $\uparrow 6.3$} & 82.7 \textcolor{ForestGreen}{\footnotesize $\uparrow 3.7$} & 82.5 \textcolor{ForestGreen}{\footnotesize $\uparrow 2.1$} & 27.1 \textcolor{ForestGreen}{\footnotesize $\uparrow 3.2$} \\
        & \checkmark & \checkmark & 90.2 \textcolor{ForestGreen}{\footnotesize $\uparrow 7.2$} & 83.4 \textcolor{ForestGreen}{\footnotesize $\uparrow 4.5$} & 93.4 \textcolor{ForestGreen}{\footnotesize $\uparrow 13.0$} & 41.7 \textcolor{ForestGreen}{\footnotesize $\uparrow 17.8$} \\
        \midrule

        \multirow{4}{*}{DINOv3~\cite{simeoni2025dinov3}} & & & 85.3 & 80.1 & 88.0 & 32.5 \\
        & \checkmark & & 89.0 \textcolor{ForestGreen}{\footnotesize $\uparrow 3.7$}  & 82.0 \textcolor{ForestGreen}{\footnotesize $\uparrow 1.9$}  & 95.0 \textcolor{ForestGreen}{\footnotesize $\uparrow 7.0$}  & 44.9 \textcolor{ForestGreen}{\footnotesize $\uparrow 12.4$}  \\
        & & \checkmark &  90.5 \textcolor{ForestGreen}{\footnotesize $\uparrow 5.2$}  & 83.9 \textcolor{ForestGreen}{\footnotesize $\uparrow 3.8$}  & 90.2 \textcolor{ForestGreen}{\footnotesize $\uparrow 2.2$}  & 39.3 \textcolor{ForestGreen}{\footnotesize $\uparrow 7.3$}  \\
        & \checkmark & \checkmark & 91.5 \textcolor{ForestGreen}{\footnotesize $\uparrow 6.2$}  & 84.7 \textcolor{ForestGreen}{\footnotesize $\uparrow 4.6$}  & 95.3 \textcolor{ForestGreen}{\footnotesize $\uparrow 7.2$}  & 44.6 \textcolor{ForestGreen}{\footnotesize $\uparrow 12.1$}  \\
        \midrule

        \multirow{4}{*}{RADIO~\cite{Ranzinger_2024_CVPR}} & & &  89.1 & 83.9 & 84.9 & 30.8 \\
        & \checkmark & & 92.1 \textcolor{ForestGreen}{\footnotesize $\uparrow 3.0$}  & 87.2 \textcolor{ForestGreen}{\footnotesize $\uparrow 3.3$}  & 95.9 \textcolor{ForestGreen}{\footnotesize $\uparrow 11.0$}  & 43.2 \textcolor{ForestGreen}{\footnotesize $\uparrow 12.6$}  \\
        & & \checkmark & 93.0 \textcolor{ForestGreen}{\footnotesize $\uparrow 3.9$}  & 88.9 \textcolor{ForestGreen}{\footnotesize $\uparrow 5.0$}  & 95.2 \textcolor{ForestGreen}{\footnotesize $\uparrow 10.3$}  & 42.8 \textcolor{ForestGreen}{\footnotesize $\uparrow 12.0$}  \\
        & \checkmark & \checkmark & 94.1 \textcolor{ForestGreen}{\footnotesize $\uparrow 5.0$}  & 87.6 \textcolor{ForestGreen}{\footnotesize $\uparrow 3.7$}  & 96.9 \textcolor{ForestGreen}{\footnotesize $\uparrow 12.0$}  & 44.3 \textcolor{ForestGreen}{\footnotesize $\uparrow 13.5$}  \\
        \bottomrule
    \end{tabular}
}
\vspace{-0.2cm}
\end{table}

\begin{table*}[]
    \centering
    \caption{Comparisons of zero-shot anomaly detection methods on industrial inspection datasets. The best performance is colored in \textcolor{red}{red} and the second best in \textcolor{blue}{blue}.}
    \label{tab:industrial}
    \vspace{-8pt}
    \resizebox{1\linewidth}{!}{
    \setlength{\tabcolsep}{1mm}
    \begin{tabular}{@{}cccccccccccc@{}}
\toprule
\multirow{2}{*}{Metric} & \multirow{2}{*}{Dataset} & SAA~\cite{cao2023segment} & WinCLIP~\cite{jeong2023winclip} & AnomalyCLIP~\cite{zhou2024anomalyclip} & AdaCLIP~\cite{cao2024adaclip} & AACLIP~\cite{ma2025aa} & Bayes-PFL~\cite{BayesFPL} & FAPrompt~\cite{zhu2025fine} & \textit{AnomalyVFM} \\ 
 & & ToC'25 & CVPR'23 & ICLR'24 & ECCV'24 & CVPR'25 & CVPR'25 & ICCV'25 &  \\ 
\midrule 
\multirow{10}{*}{\begin{tabular}[c]{@{}c@{}}Image-level\\      (AUROC, max-F1)\end{tabular}}
& MVTec AD & (63.5, 87.4) & (91.8, 92.9) & (91.6, 92.7) & (89.2, 90.6) & (90.5, 90.4) & (\textcolor{blue}{92.3}, \textcolor{blue}{93.1}) & (91.1, 92.2) & (\textcolor{red}{94.9}, \textcolor{red}{94.1}) \\ 
& VisA & (67.1, 75.9) & (78.1, 80.7) & (82.0, 80.4) & (85.8, 83.1) & (84.6, 78.8) & (\textcolor{blue}{87.0}, \textcolor{blue}{84.1}) & (82.8, 81.3) & (\textcolor{red}{93.6}, \textcolor{red}{90.1}) \\ 
& BTAD & (59.0, 89.7) & (68.2, 67.8) & (88.2, 83.8) & (88.6, 88.2) & (\textcolor{blue}{94.8}, \textcolor{red}{93.7}) & (93.2, \textcolor{blue}{91.9}) & (90.7, 88.1) & (\textcolor{red}{96.0}, 91.0) \\ 
& MPDD & (42.7, 73.9) & (61.4, 77.5) & (77.5, 80.4) & (76.0, 82.5) & (75.1, 79.8) & (\textcolor{blue}{81.2}, \textcolor{blue}{83.5}) & (76.6, 80.4) & (\textcolor{red}{85.5}, \textcolor{red}{87.8}) \\ 
& RealIAD & (51.4, 64.6) & (74.7, 69.8) & (78.7, \textcolor{blue}{80.0}) & (79.2, 73.5) & (81.3, 76.4) & (\textcolor{blue}{85.2}, 78.7) & (81.6, 75.2) & (\textcolor{red}{88.0}, \textcolor{red}{81.6}) \\ 
& KSDD & (68.6, 37.6) & (93.3, \textcolor{blue}{79.0}) & (84.5, 71.1) & (\textcolor{red}{97.1}, \textcolor{red}{90.7}) & (69.3, 57.1) & (88.2, 56.0) & (81.3, 71.1) & (92.5, 69.7) \\ 
& KSDD2 & (91.6, 67.0) & (94.2, 71.5) & (94.1, 80.0) & (95.9, \textcolor{blue}{86.7}) & (95.9, 84.4) & (\textcolor{red}{97.3}, \textcolor{red}{87.6}) & (95.6, 84.8) & (\textcolor{blue}{97.1}, 79.2) \\ 
& DAGM & (87.1, 88.8) & (91.8, 87.6) & (97.7, 90.1) & (\textcolor{blue}{99.1}, \textcolor{red}{97.5}) & (93.2, 79.4) & (97.7, 95.7) & (97.3, 89.3) & (\textcolor{red}{99.6}, \textcolor{blue}{95.8}) \\ 
& DTD & (94.4, 93.5) & (95.1, 94.1) & (93.9, 93.6) & (95.5, 94.7) & (90.4, 92.8) & (95.1, \textcolor{blue}{95.1}) & (\textcolor{blue}{95.9}, 94.7) & (\textcolor{red}{99.4}, \textcolor{red}{99.0}) \\ 
\cmidrule(l){2-10} 
& \textit{Average} & (69.5, 75.4) & (83.2, 80.1) & (87.6, 83.6) & (89.6, \textcolor{blue}{87.5}) & (86.1, 81.4) & (\textcolor{blue}{90.8}, 85.1) & (88.1, 84.1) & (\textcolor{red}{94.1}, \textcolor{red}{87.6}) \\ 
\midrule 
\multirow{10}{*}{\begin{tabular}[c]{@{}c@{}}Pixel-level\\      (AUROC, max-F1)\end{tabular}}
& MVTec AD & (75.5, 38.1) & (88.7, 43.4) & (91.1, 39.1) & (88.7, 43.4) & (91.4, \textcolor{blue}{46.4}) & (\textcolor{blue}{91.8}, \textcolor{red}{49.0}) & (90.8, 39.3) & (\textcolor{red}{92.7}, 45.2) \\ 
& VisA & (76.5, 31.6) & (95.5, \textcolor{red}{37.7}) & (95.5, 28.3) & (95.5, \textcolor{red}{37.7}) & (94.8, 30.2) & (\textcolor{blue}{95.6}, 34.3) & (\textcolor{blue}{95.6}, 27.6) & (\textcolor{red}{96.2}, 31.2) \\ 
& BTAD & (65.8, 14.8) & (92.1, 51.7) & (94.2, 49.7) & (92.1, 51.7) & (\textcolor{red}{97.3}, \textcolor{red}{55.1}) & (93.9, 52.0) & (\textcolor{blue}{95.8}, \textcolor{blue}{52.6}) & (92.3, 49.7) \\ 
& MPDD & (81.7, 18.9) & (96.1, 34.9) & (96.5, 34.2) & (96.1, 32.8) & (96.7, 30.0) & (\textcolor{red}{97.8}, \textcolor{blue}{35.0}) & (95.5, 31.9) & (\textcolor{blue}{97.0}, \textcolor{red}{38.1}) \\ 
& RealIAD & (73.5, 4.5) & (87.2, 10.8) & (96.3, 39.0) & (\textcolor{red}{97.2}, \textcolor{red}{43.0}) & (96.2, 40.2) & (\textcolor{red}{97.2}, \textcolor{blue}{41.2}) & (96.2, 38.3) & (96.4, 40.4) \\ 
& KSDD & (78.8, 6.6) & (\textcolor{blue}{97.7}, \textcolor{red}{54.5}) & (90.6, 42.5) & (\textcolor{blue}{97.7}, \textcolor{red}{54.5}) & (87.1, 28.0) & (96.5, 6.6) & (93.1, 47.2) & (\textcolor{red}{99.0}, 10.1) \\ 
& KSDD2 & (79.9, \textcolor{blue}{63.4}) & (94.4, 23.9) & (98.5, 59.8) & (98.5, \textcolor{red}{67.0}) & (\textcolor{red}{99.5}, \textcolor{blue}{63.4}) & (97.0, 62.0) & (99.1, 60.4) & (\textcolor{blue}{99.3}, 55.9) \\ 
& DAGM & (91.5, 57.5) & (91.5, 57.5) & (95.6, 58.9) & (91.5, 57.5) & (96.2, 53.3) & (95.9, 49.8) & (\textcolor{blue}{98.6}, \textcolor{blue}{60.2}) & (\textcolor{red}{99.4}, \textcolor{red}{61.3}) \\ 
& DTD & (97.9, \textcolor{red}{71.6}) & (97.9, \textcolor{red}{71.6}) & (97.9, 62.2) & (97.9, \textcolor{red}{71.6}) & (95.8, 59.6) & (\textcolor{blue}{98.4}, 65.2) & (98.1, 61.9) & (\textcolor{red}{99.4}, 66.5) \\ 
\cmidrule(l){2-10} 
& \textit{Average} & (80.1, 34.1) & (93.5, 42.9) & (95.1, 46.0) & (95.0, \textcolor{red}{51.0}) & (95.0, 45.1) & (\textcolor{blue}{96.0}, 43.9) & (95.9, \textcolor{blue}{46.6}) & (\textcolor{red}{96.9}, 44.3) \\ 
\bottomrule
\end{tabular}}
\vspace{-0.1cm}
\end{table*}

\begin{table*}[]
    \centering
    \caption{Comparisons of zero-shot anomaly detection methods on medical datasets. $\dagger$ - AdaCLIP is also trained with auxiliary medical datasets. Other methods are not.}
    \label{tab:medical}
    \vspace{-8pt}
    \resizebox{1\linewidth}{!}{
    \setlength{\tabcolsep}{1mm}
    \begin{tabular}{@{}cccccccccccc@{}}
\toprule
\multirow{2}{*}{Metric} & \multirow{2}{*}{Dataset} & SAA~\cite{cao2023segment} & WinCLIP~\cite{jeong2023winclip} & AnomalyCLIP~\cite{zhou2024anomalyclip} & AdaCLIP$^\dagger$~\cite{cao2024adaclip} & AACLIP~\cite{ma2025aa} & Bayes-PFL~\cite{BayesFPL} & FAPrompt~\cite{zhu2025fine} & \textit{AnomalyVFM} \\ 
 & & ToC'25 & CVPR'23 & ICLR'24 & ECCV'24 & CVPR'25 & CVPR'25 & ICCV'25 &  \\ 
\midrule 
\multirow{4}{*}{\begin{tabular}[c]{@{}c@{}}Image-level\\      (AUROC, max-F1)\end{tabular}}
& HeadCT & (46.8, 68.0) & (84.1, 79.8) & (93.0, 88.4) & (91.4, 85.2) & (\textcolor{red}{96.9}, \textcolor{red}{93.1}) & (92.6, 86.3) & (93.0, 88.2) & (\textcolor{blue}{94.8}, \textcolor{blue}{90.5}) \\ 
& BrainMRI & (34.4, 76.7) & (89.9, 86.9) & (90.0, 86.5) & (94.8, 91.2) & (80.2, 91.5) & (\textcolor{blue}{95.2}, \textcolor{red}{94.4}) & (\textcolor{red}{95.5}, \textcolor{blue}{93.2}) & (92.9, 92.5) \\ 
& BR35H & (33.2, 67.3) & (81.6, 74.4) & (94.2, 86.8) & (\textcolor{red}{97.7}, \textcolor{blue}{92.4}) & (95.4, 90.2) & (\textcolor{blue}{97.0}, \textcolor{red}{93.2}) & (96.6, 90.3) & (94.4, 90.2) \\ 
\cmidrule(l){2-10} 
& \textit{Average} & (38.1, 70.7) & (85.2, 80.4) & (92.4, 87.2) & (94.6, 89.6) & (90.8, \textcolor{red}{91.6}) & (\textcolor{blue}{94.9}, \textcolor{blue}{91.3}) & (\textcolor{red}{95.0}, 90.6) & (94.0, 91.1) \\ 
\midrule 
\multirow{7}{*}{\begin{tabular}[c]{@{}c@{}}Pixel-level\\      (AUROC, max-F1)\end{tabular}}
& ISIC & (83.8, 74.2) & (83.3, 64.1) & (89.4, 71.6) & (89.3, 71.4) & (\textcolor{red}{94.6}, 80.4) & (92.3, 76.8) & (90.1, 72.0) & (90.8, 74.4) \\ 
& ClinicDB & (66.2, 29.1) & (74.3, 30.7) & (82.9, 42.4) & (84.4, \textcolor{red}{58.2}) & (\textcolor{blue}{89.6}, 54.1) & (\textcolor{blue}{89.6}, 51.7) & (83.2, 43.4) & (\textcolor{red}{92.0}, \textcolor{blue}{57.6}) \\ 
& ColonDB & (71.8, 31.5) & (61.2, 19.6) & (81.9, 37.5) & (\textcolor{red}{90.4}, \textcolor{red}{58.2}) & (84.1, 38.1) & (82.1, 39.2) & (84.1, 38.8) & (85.6, 42.8) \\ 
& Kvasir & (86.2, \textcolor{blue}{65.9}) & (38.6, 27.0) & (79.0, 46.2) & (\textcolor{red}{95.0}, \textcolor{red}{77.1}) & (87.3, 57.3) & (85.3, 54.8) & (81.6, 48.8) & (\textcolor{blue}{90.6}, 63.2) \\ 
& Endo & (79.4, 51.6) & (43.7, 25.3) & (84.2, 50.3) & (\textcolor{red}{96.6}, \textcolor{red}{80.1}) & (90.2, 59.7) & (89.2, 57.9) & (86.4, 52.7) & (\textcolor{blue}{92.2}, \textcolor{blue}{64.4}) \\ 
& TN3K & (66.8, 32.6) & (67.2, 30.0) & (81.4, 47.8) & (77.2, 41.9) & (80.5, 43.0) & (\textcolor{blue}{85.4}, 42.4) & (84.4, \textcolor{blue}{49.2}) & (\textcolor{red}{89.0}, \textcolor{red}{55.6}) \\ 
\cmidrule(l){2-10} 
& \textit{Average} & (75.7, 47.5) & (61.4, 32.8) & (83.1, 49.3) & (\textcolor{blue}{88.8}, \textcolor{red}{64.5}) & (87.7, 55.4) & (87.3, 53.8) & (85.0, 50.8) & (\textcolor{red}{90.0}, \textcolor{blue}{59.7}) \\ 
\bottomrule
\end{tabular}}
\vspace{-0.3cm}
\end{table*}

\subsection{Datasets}
\label{sec:dat}

We evaluate AnomalyVFM on 9 industrial and 9 medical anomaly detection datasets as standard in other zero-shot methods~\cite{deng2022anomaly,cao2024adaclip}. For industrial anomaly detection, MVTec AD~\cite{bergmann2019mvtec}, VisA~\cite{visa2022}, BTAD~\cite{mishra21-btad}, MPDD~\cite{mpdd}, Real-IAD~\cite{wang2024real}, KSDD~\cite{ksdd}, KSDD2~\cite{KSDD2}, DAGM~\cite{dagm}, and DTD-Synthetic~\cite{dtdsynth} were used, and for medical anomaly detection, HeadCT~\cite{salehi2021multiresolution}, BrainMRI~\cite{kanade2015brain}, BR35H~\cite{hamada2020br35h}, ISIC~\cite{codella2018skin}, ClinicDB~\cite{bernal2015wm}, ColonDB~\cite{tajbakhsh2015automated}, Kvasir~\cite{jha2019kvasir}, Endo~\cite{hicks2021endotect} and TN3K~\cite{gong2021multi} were used. The evaluation metrics follow AdaCLIP~\cite{cao2024adaclip}, where the AUROC and F1-max are used for image-level anomaly detection, and the pixel-wise AUROC and the pixel-wise F1-max are used for anomaly localisation. We compare AnomalyVFM to recent state-of-the-art approaches that are trained on auxiliary data. More specifically, when evaluated on the MVTec AD~\cite{bergmann2019mvtec} dataset, the recent zero-shot AD methods are trained on the VisA~\cite{visa2022} test set and are trained on the MVTec AD test set when evaluated on other datasets. In contrast, AnomalyVFM is trained solely on automatically generated data.

\subsection{Implementation Details}
In the data generation pipeline, the FLUX~\cite{flux2024} conditional image generation model is used. The $(w_{min},w_{max})$ and $(h_{min}, h_{max})$ are set to $(50,350)$ in all experiments, when generating images of dimension $1024 \times 1024$. The filtering threshold, $T$, is set to 0.3 in all experiments. A synthetic dataset of 10,000 images was generated for all experiments, unless stated otherwise. More details about the generated dataset are provided in the Supplementary Material.

Zero-shot anomaly detection training is performed on generated data. AnomalyVFM is trained for $500$ iterations with a batch size of $32$ using the AdamW optimiser and a learning rate of $10^{-4}$. The RADIOv2.5~\cite{Ranzinger_2024_CVPR} ViT-L with a patch-size of $16$ is used as the backbone for most experiments. Since RADIO has been trained on multiple resolutions, the input images are resized to $768 \times 768$ for training and evaluation. The confidence parameter $\alpha$ in Equation~\ref{eq:conf} is set to $0.1$ in all experiments.

\subsection{Generalisation of the proposed framework}

First, we evaluate our contribution across a set of diverse VFMs to verify our claims. More specifically, we use DINOv2~\cite{oquab2023dinov2}, DINOv3~\cite{simeoni2025dinov3} and RADIO~\cite{Ranzinger_2024_CVPR}. We evaluate each VFM in 4 different settings to verify our contribution. We modulate two settings: the training dataset and the adaptation strategy. For the dataset, we either follow the standard practice~\cite{cao2024adaclip,BayesFPL,zhou2024anomalyclip} of training on the test set of MVTec AD (the model is trained on the test set of VisA when evaluated on MVTec AD) or the dataset generated with the proposed synthetic generation procedure. For the adaptation strategy, we either train a simple decoder and leave the internal representations unchanged or we employ the proposed adaptation strategy. To demonstrate generalisation, we evaluated our model on nine industrial datasets described in Section~\ref{sec:dat}. The results can be seen in Table~\ref{tab:contr}. All Chosen VFM achieve a significant improvement in performance in both detection and localisation, showcasing the generality and the strength of the contribution. On average, the image-level AUROC is improved by $6.1$ p.\ p.\ and the pixel-level AUROC is improved by $10.7$ p.\ p. This is also visually represented in Figure~\ref{fig:main}, where it can be seen that all of the VFMs match the performance of current methods.

\subsection{Comparison to zero-shot methods}

\begin{figure*}
    \centering
    \includegraphics[width=1.0\linewidth]{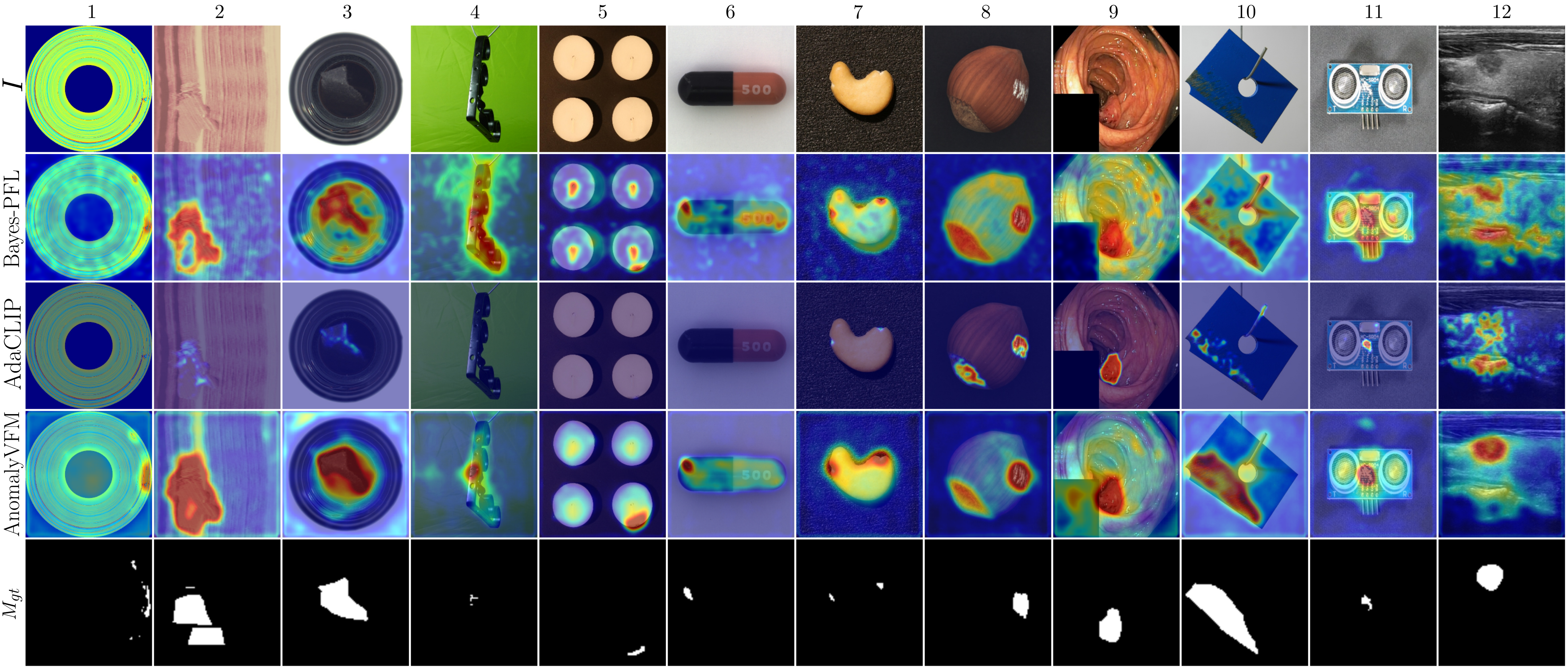}
    \caption{Qualitative comparison of the anomaly segmentation masks produced by AnomalyVFM and two other best-performing methods. In the first row, the image is shown. In the next three rows, the anomaly segmentations produced by Bayes-PFL~\cite{BayesFPL}, AdaCLIP~\cite{cao2024adaclip} and AnomalyVFM are depicted, and in the last row, the ground truth mask is depicted.
    }
    \label{fig:qualitative}
    \vspace{-0.2cm}
\end{figure*}

\noindent \textbf{Quantitative results} In Table \ref{tab:industrial}, the comparison of AnomalyVFM to the state-of-the-art zero-shot anomaly detection methods on industrial datasets is shown. AnomalyVFM outperforms the state-of-the-art considerably in both anomaly detection and anomaly localisation. More specifically, it outperforms the next best method (Bayes-PFL) in terms of image-level AUROC by a substantial 3.3 percentage points (p.\ p.). AnomalyVFM also slightly improves the results in terms of image-level F1-Max. Additionally, AnomalyVFM significantly improves results on the widely used MVTec AD, VisA, and Real IAD, achieving results close to those of full-shot methods, reiterating the contribution of our model.

In terms of anomaly localisation, AnomalyVFM also improves upon previous methods in terms of pixel-level AUROC and achieves competitive results in terms of pixel-level F1-Max. More specifically, AnomalyVFM improves previous methods by $0.9$ p.\ p.\ in terms of pixel-level AUROC. In terms of pixel-level F1-Max, it trails behind AdaCLIP~\cite{cao2024adaclip}, which achieves lower scores in terms of pixel-level AUROC. 

The results for zero-shot anomaly detection and localisation in the medical domain are presented in Table~\ref{tab:medical}. AnomalyVFM achieves competitive results in terms of detection and improves previous methods in terms of localisation scores. In terms of pixel-level AUROC, AnomalyVFM improves previous methods by 1.2 p.\ p. More importantly, the results demonstrate the generalisation of AnomalyVFM to the medical domain, despite not being finetuned on any medical data.

\noindent \textbf{Qualitative results} Qualitative examples can be seen in Figure~\ref{fig:qualitative}. AnomalyVFM produces sharper anomaly masks in comparison to Bayes-PFL and is able to localise anomalies even in cases where AdaCLIP fails. Additionally, it is able to detect both small defects (Columns 4 and 5) and larger defects (Columns 2, 3 and 10). Additionally, it successfully detects medical defects (Columns 9 and 12).

\subsection{Comparison to few-shot methods}
To verify the effectiveness of AnomalyVFM as a backbone, we have fine-tuned the zero-shot model for an additional 50 iterations using a few normal samples. We have chosen MVTec AD~\cite{bergmann2019mvtec} and VisA~\cite{visa2022} as our evaluation datasets due to their widespread use. The results can be seen in Table~\ref{tab:few_shot}. AnomalyVFM achieves the highest image-level AUROC in all settings on MVTec AD and in the 1-shot setting on VisA. 

Remarkably, despite being designed for the zero-shot regime, AnomalyVFM matches or even surpasses the performance of recent few-shot methods such as INP-Former \cite{luo2025exploring}, without any architecture changes and with minimal fine-tuning.
These results highlight the robustness and transferability of AnomalyVFM, underscoring its potential as a strong and versatile backbone for future anomaly detection research.

\begin{table}[]
\caption{Comparison to few-shot methods on MVTec AD and VisA benchmarks. Results are in image-level AUROC. The best results are marked in bold.}
\centering
\vspace{-8pt}
\resizebox{1.0\columnwidth}{!}{
\setlength{\tabcolsep}{1mm}
\begin{tabular}{lcccccc}
\toprule
\multicolumn{1}{l}{\multirow{2}{*}{Method}} &  \multicolumn{3}{c}{MVTec AD}        & \multicolumn{3}{c}{VisA}       \\
\cmidrule(r){2-4} \cmidrule(r){5-7}
& 1-shot   & 2-shot    & 4-shot    & 1-shot   & 2-shot   & 4-shot   \\ \midrule
PatchCore~\cite{roth2022towards}{\scriptsize CVPR'22}  & 83.4 & 86.3 & 88.8 & 79.9 & 81.6 & 85.3 \\
WinCLIP+~\cite{jeong2023winclip}{\scriptsize CVPR'23}  & 93.1 & 94.4  & 95.2  & 83.8 & 84.6 & 87.3 \\
PromptAD~\cite{li2024promptad}{\scriptsize CVPR'24}     & 94.6 & 95.7 & 96.6 & 86.9 & 88.3 & 89.1 \\ 
INP-Former~\cite{luo2025exploring}{\scriptsize CVPR'25}& 96.1 & 97.0 & 97.6 & 91.4 & \textbf{94.6} & \textbf{96.4} \\ 
\midrule
\multicolumn{1}{l}{AnomalyVFM}  & \textbf{97.2} & \textbf{97.9} & \textbf{98.2}& \textbf{93.8} & 94.2 & 94.5 \\

\bottomrule
\end{tabular}
}
\label{tab:few_shot}
\vspace{-0.3cm}
\end{table}

\begin{table*}[t]
\centering
\caption{Ablation of the anomaly detection method components.}
\label{tab:model_ablation}
\vspace{-8pt}
\resizebox{0.65\textwidth}{!}{
\begin{tabular}{llcccc} 
         \toprule
         \multirow{2}{*}{\textbf{Group}} & \multirow{2}{*}{\textbf{Condition}}  & \multicolumn{2}{c}{Image-level} & \multicolumn{2}{c}{Pixel-level}  \\
          \cmidrule(r){3-4} \cmidrule(r){5-6}&   & AUROC & $F_1$-Max & AUROC & $F_1$-Max \\
        \midrule
        \multirow{4}{*}{\textit{Image Generation}} & FLUX~\cite{flux2024} $\rightarrow$ QWEN-Image~\cite{wu2025qwen} & \textcolor{gray}{-0.1} & \textcolor{gray}{+0.1} & \textcolor{gray}{-0.5} & \textcolor{gray}{-0.2}  \\
        & FLUX~\cite{flux2024} $\rightarrow$ WAN~\cite{wan2025wan} & \textcolor{gray}{-0.4} & \textcolor{gray}{-0.8} & \textcolor{gray}{-2.1} & \textcolor{gray}{-1.0}  \\
        
        & No Filtering & \textcolor{gray}{-3.8} & \textcolor{gray}{-2.7} & \textcolor{gray}{-14.6} & \textcolor{gray}{-18.7}  \\
        & No Foreground Selection & \textcolor{gray}{-1.4} & \textcolor{gray}{-0.8} & \textcolor{gray}{-5.8} & \textcolor{gray}{-12.3}  \\
        \midrule
        \multirow{3}{*}{\textit{Module Ablation}} & No Confidence Loss & \textcolor{gray}{-0.6} & \textcolor{gray}{-0.1} & \textcolor{gray}{-2.0} & \textcolor{gray}{-6.3}  \\
        & LoRA~\cite{hu2022lora} $\rightarrow$ AdaLN~\cite{lian2022scaling} & \textcolor{gray}{-0.7} & \textcolor{gray}{-1.0} & \textcolor{gray}{-2.9} & \textcolor{gray}{-1.9}  \\
        & LoRA~\cite{hu2022lora} $\rightarrow$ VPT~\cite{jia2022visual} & \textcolor{gray}{-1.0} & \textcolor{gray}{+1.2} & \textcolor{gray}{-0.5} & \textcolor{gray}{+4.0}  \\ \midrule
        \textit{AnomalyVFM} & & 94.1 & 87.6 & 96.9 & 44.3 \\
        \bottomrule
    \end{tabular}
}
\vspace{-0.3cm}
\end{table*}

\section{Ablation study}

Ablation experiments validating the individual contributions of AnomalyVFM are performed on 9 industrial datasets presented in the Section~\ref{sec:dat}. Results are shown in Table~\ref{tab:model_ablation}. Additional experiments are presented in the Supplementary Material.

\noindent \textbf{Image Generation Model} FLUX~\cite{flux2024} is used as the default image generation model in our experiments. To verify the importance of this choice, we replaced it with two recent generative models: QWEN-Image~\cite{wu2025qwen} and WAN~\cite{wan2025wan}. This leads to a very slight decrease in performance: $0.1$ p.\ p.\ and $0.4$ p.\ p.\ in image-level AUROC, respectively and 
$0.5$ p.\ p.\ and $2.1$ p.\ p.\ in pixel-level AUROC, respectively. This shows that our generation pipeline is robust to this choice.

\noindent \textbf{Dataset Filtering} To measure the importance of verifying that the generated images actually do contain anomalies, we have omitted the dataset filtering step. This leads to a decrease in image-level AUROC for a significant $3.8$ p.\ p.\ and a decrease in pixel-level AUROC for $14.6$ p.\ p. This showcases both the problems with current image generation models and the necessity of having clean data.

\noindent \textbf{Anomaly Location Importance} In our pipeline, the anomaly region is selected by sampling a rectangle $R$ on the foreground $M_{fg}$ produced by an external model. To verify the importance of this step, we set $M_{fg}$ to be equal to the whole image. This means that sometimes the generated images contain a defect in the background, so the model is trained to focus not only on the object but also on the background. This leads to a decrease of $1.4$ p.\ p.\ in image-level AUROC and $5.8$ p.\ p.\ in pixel-level AUROC. This highlights the importance of selecting the inpainting location intelligently.

\noindent \textbf{Confidence Loss} To show the importance of the introduced confidence loss, we have retrained the model without it. This has led to a slight decrease in both image-level and pixel-level AUROC ($0.6$ p.\ p.\ and $2.0$ p.\ p.\, respectively). This reiterates the importance of this loss for optimal performance. 

\noindent \textbf{Adapter Architecture} To even further show the generality of our framework, LoRA~\cite{hu2022lora} adapters were exchanged with two other Parameter Efficient Techniques, AdaLN~\cite{lian2022scaling} and VPT~\cite{jia2022visual}. This has led to a decrease in performance of $0.7$ p.\ p.\ and $1.0$ p.\ p.\ in image-level AUROC and $2.9$ p.\ p.\ and $0.5$ p.\ p.\ in pixel-level AUROC, respectively. All of these results are still significantly above SOTA and show the robustness of our framework to this choice. Additionally, it shows possible extensions to newer VFMs, which might have different architectures.

\noindent \textbf{Inference Speed and Computational Complexity} The inference speed can be seen in Table~\ref{tb:inf_speed}. The protocol from EfficientAD~\cite {batzner2024efficientad} was used to calculate them. AnomalyVFM is significantly faster than its main competitors. AnomalyVFM requires approximately 2 hours to train on a single A100 GPU and has $345.8$ million parameters. Out of these $35.4$ million are trainable.

\begin{table}[t]
\centering
\caption{Results for average inference time of a single sample with NVIDIA A100 GPU. Inference times are reported in milliseconds.}
\vspace{-8pt}
\resizebox{\columnwidth}{!}{
\begin{tabular}{lccc} \toprule
\textbf{Method} & Bayes-PFL~\cite{BayesFPL} & AdaCLIP~\cite{cao2024adaclip}  & \textit{AnomalyVFM} \\ \midrule
Inference [ms] & 208.5 & 82.4 & \textbf{20.5} \\ \bottomrule
\end{tabular}
}

\vspace{-0.4cm}
\label{tb:inf_speed}
\end{table}

\noindent \textbf{Limitations} At present, the main bottleneck lies in the image generation stage, which takes approximately one day on an A100 GPU, whereas model training requires only about two hours. Additional discussion of these limitations and related analyses can be found in the Supplementary.

\section{Conclusion}

We present AnomalyVFM, a practical and model-agnostic framework that transforms any pretrained Vision Foundation Model into a strong zero-shot anomaly detector. Unlike prior approaches that rely on high-level concept knowledge from vision–language models, AnomalyVFM leverages the rich visual representations of VFMs and enhances them through two key innovations. First, we introduced a three-stage synthetic dataset generator that produces diverse and realistic training samples, capturing a broad range of object categories and defect types. Second, we designed a parameter-efficient adaptation strategy that inserts low-rank adapters throughout the backbone and employs a confidence-weighted loss to refine the model’s representations with minimal parameters and robust supervision.

Together, these components allow VFMs to generalise to unseen object classes and outperform existing VLM-based methods in the zero-shot regime. More specifically, we achieve an average image-level AUROC of 94.1\% across 9 diverse industrial datasets, improving upon previous methods by a significant 3.3. percentage points. Additionally, we demonstrate the effectiveness of AnomalyVFM as a potent backbone by finetuning it on a few normal samples without any bells and whistles. With this, AnomalyVFM achieves a performance comparable to SOTA in the few-shot regime.

Looking ahead, further efforts to improve defect realism and resulting labels are a good avenue for future research. Additionally, integrating depth data via monodepth foundational models, such as Marigold~\cite{ke2024repurposing}, could be used to enable zero-shot RGBD anomaly detection. Most importantly, the results also indicate that AnomalyVFM could be used as a backbone for future few-shot and full-shot models.

\noindent \textbf{Acknowledgements} This work was in part supported by the ARIS research projects MUXAD (J2-60055) and AI4Science (GC-0001), research programme P2-0214 and the supercomputing network SLING (ARNES, EuroHPC Vega).
{
    \small
    \bibliographystyle{ieeenat_fullname}
    \bibliography{main}
}

\clearpage
\setcounter{page}{1}
\setcounter{footnote}{0}
\setcounter{table}{0}
\setcounter{figure}{0}
\maketitlesupplementary

\appendix

In this Appendix, we provide extensive additional details and supporting information that extend beyond the scope of the main manuscript. The Appendix is organised as follows: 
\begin{itemize}
    \item \textbf{Limitations} in Section~\ref{a:limit_social}.
    \item \textbf{Discussion about the Dataset Generation Phase} in Section~\ref{a:img_gen}.
    \item \textbf{Results of competing methods when trained on the synthetic dataset and a discussion about them} in Section~\ref{a:comp_meth}.
    \item \textbf{Extended synthetic dataset details} in Section~\ref{a:syn_data}.
    \item \textbf{Extended ablation studies} in Section~\ref{a:abl}.
    \item \textbf{Additional qualitative results} in Section~\ref{a:qual}.
    \item \textbf{Data generation data} in Section~\ref{a:gen_data}.
    
\end{itemize}

\section{Limitations}
\label{a:limit_social}

The main limitation currently is the time required to generate the synthetic dataset, which takes approximately one day on an A100 GPU, whereas model training requires only about two hours. While this represents a lot of time, it is a one-time investment, and the same dataset can be used for every VFM. With the improvements to the generation speed of current image generation models, we expect this time to drop even further. In Section~\ref{a:abl}, we also conducted additional experiments, demonstrating that good performance can be achieved with fewer than 10,000 images, meaning the generation phase can be shorter if needed. 

Additionally, while AnomalyVFM performs well on medical datasets, its performance could be further improved. In our preliminary attempts, the pretrained image generation models~\cite{flux2024,wu2025qwen,wan2025wan} failed to output realistic medical images suitable for zero-shot anomaly detection training. While this was not needed for industrial anomaly detection, fine-tuning the image generator on an auxiliary medical imaging dataset may enable the image-generation model to output data of suitable quality.

\begin{figure}[t]
    \centering
    \includegraphics[width=1.0\columnwidth]{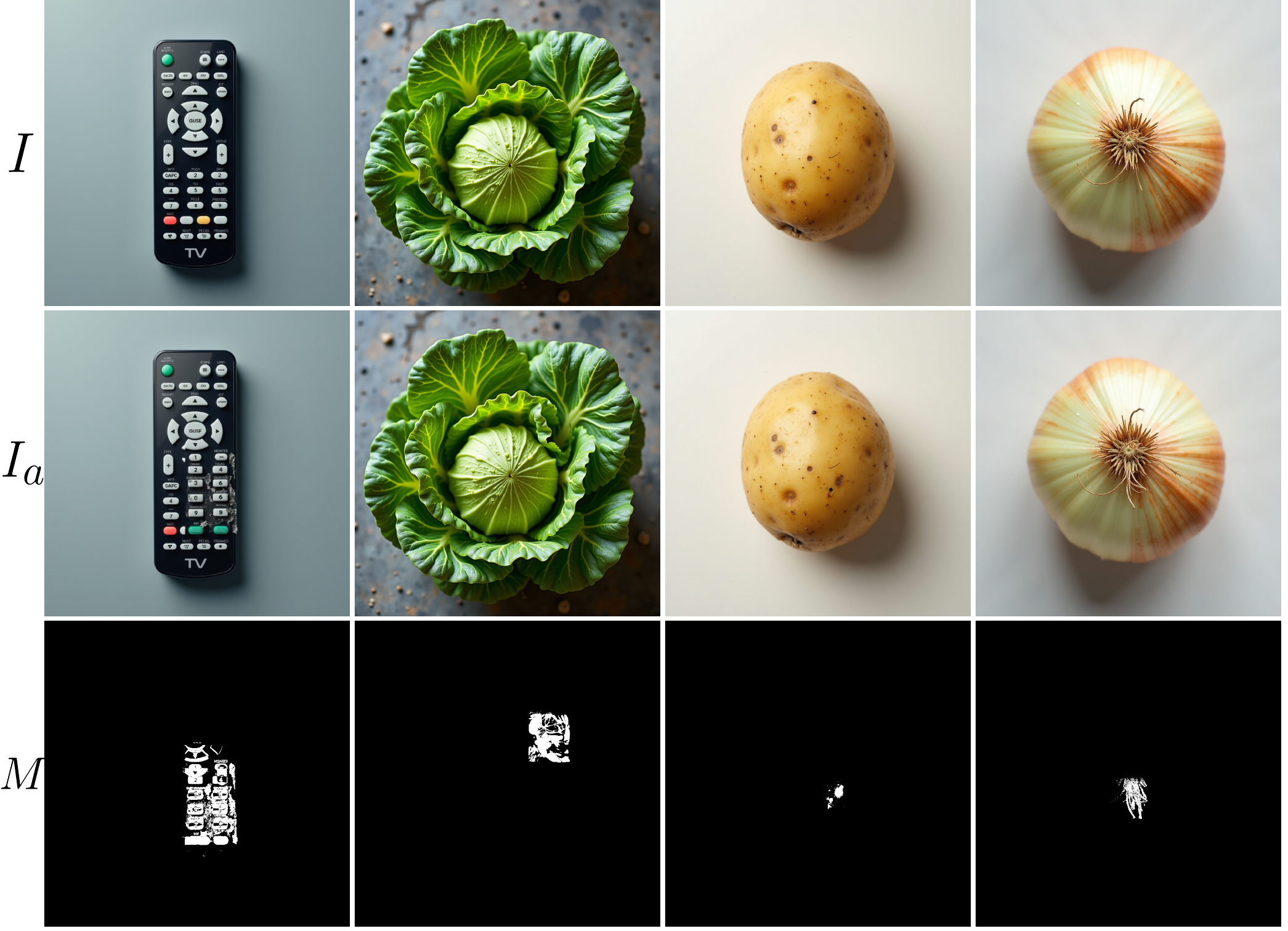}
    \caption{Failure Cases in Image Generation Process
    }
    \label{fig:failure}
\end{figure}

\begin{table}[]
\caption{Comparison of performance of competing methods when trained on the proposed synthetic dataset versus when using the default datasets. SD stands for Synthetic Dataset}
\centering
\vspace{-8pt}
\resizebox{1.0\columnwidth}{!}{
\setlength{\tabcolsep}{1mm}
\begin{tabular}{lccccc}
\toprule
\multicolumn{1}{l}{\multirow{2}{*}{Method}} & \multirow{2}{*}{SD} &  \multicolumn{2}{c}{Image-level}        & \multicolumn{2}{c}{Pixel-level}       \\
\cmidrule(r){3-4} \cmidrule(r){5-6}
& & AUROC & F1-Max & AUROC & F1-Max \\ \midrule
\multirow{2}{*}{AACLIP~\cite{ma2025aa}} & & 86.1 & 81.4 & 95.0 & 45.1 \\
& \checkmark & 85.6 \textcolor{BrickRed}{\footnotesize $\downarrow 0.5$} & 81.1 \textcolor{BrickRed}{\footnotesize $\downarrow 0.3$} & 93.5 \textcolor{BrickRed}{\footnotesize $\downarrow 1.5$} & 44.2 \textcolor{BrickRed}{\footnotesize $\downarrow 0.9$} \\ \midrule

\multirow{2}{*}{AnomalyCLIP~\cite{zhou2024anomalyclip}} & & 87.6 & 83.6 & 95.1 & 46.0 \\
& \checkmark & 87.5 \textcolor{BrickRed}{\footnotesize $\downarrow 0.1$} & 82.5 \textcolor{BrickRed}{\footnotesize $\downarrow 1.1$} & 95.3 \textcolor{ForestGreen}{\footnotesize $\uparrow 0.3$} & 45.8 \textcolor{BrickRed}{\footnotesize $\downarrow 0.2$} \\ \midrule

\multirow{2}{*}{FAPrompt~\cite{zhu2025fine}} & & 88.1 & 84.1 & 95.9 & 46.6\\
& \checkmark & 88.5 \textcolor{ForestGreen}{\footnotesize $\uparrow 0.4$} & 84.4 \textcolor{ForestGreen}{\footnotesize $\uparrow 0.3$} & 96.2 \textcolor{ForestGreen}{\footnotesize $\uparrow 0.3$} & 45.9 \textcolor{BrickRed}{\footnotesize $\downarrow 0.7$} \\ \midrule

\multirow{2}{*}{AdaCLIP~\cite{cao2024adaclip}} & & 89.6 & 87.5 & 95.0 & 51.0\\
& \checkmark & 87.1 \textcolor{BrickRed}{\footnotesize $\downarrow 2.5$} & 84.9 \textcolor{BrickRed}{\footnotesize $\downarrow 2.6$}& 92.9 \textcolor{BrickRed}{\footnotesize $\downarrow 2.1$}& 47.9 \textcolor{BrickRed}{\footnotesize $\downarrow 3.1$}\\  \midrule

\multirow{2}{*}{Bayes-PFL~\cite{BayesFPL}} & & 90.8 & 85.1 & 96.0 & 43.9\\
& \checkmark & 91.2 \textcolor{ForestGreen}{\footnotesize $\uparrow 0.4$} & 85.6 \textcolor{ForestGreen}{\footnotesize $\uparrow 0.5$}& 96.1 \textcolor{ForestGreen}{\footnotesize $\uparrow 0.1$}& 43.8 \textcolor{BrickRed}{\footnotesize $\downarrow 0.1$}\\

\bottomrule
\end{tabular}
}
\label{tab:comp_syn_data}
\vspace{-10pt}
\end{table}

\section{Discussion about dataset generation phase}
\label{a:img_gen}

While our synthetic dataset generation works well, it could be further improved. More specifically, the anomaly mask estimation and image filtering could be further improved. Although the dataset filtering is quite robust, some images without anomalies still pass through. Some examples of this can be seen in Figure~\ref{fig:failure}. A trained AnomalyVFM could be used to further filter the data and improve the data quality even further. On top of that, the amount and the content of $\texttt{[Object]}$ tags could be improved. Based on the experiment in Section~\ref{a:abl}, we hypothesise that this would improve the performance even further. We have, however, left this for future work.

To ensure that \textbf{no data leakage} occurred during the generation phase, we manually reviewed the $\texttt{[Object]}$ tags and excluded any tags that were included in the evaluation test sets. We have left $\texttt{[Anomaly]}$ and $\texttt{[Texture]}$ as they were generated, as these represent more general concepts.

\section{Training Competing methods with the proposed synthetic dataset}
\label{a:comp_meth}

\begin{table}[]
    \centering
    \caption{Dataset Statistics for the generated dataset}
    \vspace{-8pt}
    \begin{tabular}{lc}
        \toprule
         Dataset Statistic & Value \\ \midrule
         No. of images & 10,000 \\
         No. of different objects & 100 \\
         No. of different backgrounds & 50 \\
         No. of different anomalies & 204 \\
         No. of object background combinations & 4,596 \\
         Avg. Anomalous Area & 2.52\% \\
         Min. Anomalous Area & 0.28\% \\
         Max. Anomalous Area & 11.24\%\\
         \bottomrule
    \end{tabular}
    
    \label{tab:syn_data}
\end{table}

\begin{table*}[t]
\centering
\caption{Additional ablations of the anomaly detection method components.}
\label{tab:abl_extra}
\vspace{-8pt}
\resizebox{0.75\textwidth}{!}{
\begin{tabular}{llcccc} 
         \toprule
         \multirow{2}{*}{\textbf{Group}} & \multirow{2}{*}{\textbf{Condition}}  & \multicolumn{2}{c}{Image-level} & \multicolumn{2}{c}{Pixel-level}  \\
          \cmidrule(r){3-4} \cmidrule(r){5-6}&   & AUROC & $F_1$-Max & AUROC & $F_1$-Max \\
        \midrule
        \multirow{7}{*}{\textit{Module Ablation}} & ViT-L $\rightarrow$ ViT-B & \textcolor{gray}{-1.8} & \textcolor{gray}{-1.9} & \textcolor{gray}{-1.2} & \textcolor{gray}{-2.5} \\
        & ViT-L $\rightarrow$ ViT-H & \textcolor{gray}{-0.6} & \textcolor{gray}{-0.4} & \textcolor{gray}{-0.8} & \textcolor{gray}{-2.3} \\
        & LoRA Rank 64 $\rightarrow$ 32 & \textcolor{gray}{-0.3} & \textcolor{gray}{0.0} & \textcolor{gray}{-0.3} & \textcolor{gray}{+0.4}  \\
        & LoRA Rank 64 $\rightarrow$ 128 & \textcolor{gray}{0.0} & \textcolor{gray}{+0.1} & \textcolor{gray}{-0.3} & \textcolor{gray}{-0.2}  \\
        & LoRA Positions: QKV and Proj & \textcolor{gray}{-0.1} & \textcolor{gray}{+0.2} & \textcolor{gray}{-0.2} & \textcolor{gray}{-0.2}\\
        & LoRA Positions: All Norm Layers & \textcolor{gray}{-0.1} & \textcolor{gray}{-0.4} & \textcolor{gray}{-0.4} & \textcolor{gray}{-1.1}\\
        & LoRA Positions: All Linear Layers & \textcolor{gray}{-0.3} & \textcolor{gray}{-1.3} & \textcolor{gray}{-0.1} & \textcolor{gray}{-0.5}\\
        
        \midrule
        \textit{AnomalyVFM} & LoRA Positions: QV and Proj & 94.1 & 87.6 & 96.9 & 44.3 \\
        \bottomrule
    \end{tabular}
}
\vspace{-0.3cm}
\end{table*}

To demonstrate that the diversity of the datasets is not problematic for VLM-based methods, we retrained them using the proposed synthetic dataset. The results can be seen in Table~\ref{tab:comp_syn_data}. While it does help for some methods, it does not significantly alter the results. This indicates that VLM methods do not suffer from the same problem of inadequate data diversity as VFMs.

\section{Synthetic Dataset Details}
\label{a:syn_data}

\begin{figure}[t]
    \centering
    \includegraphics[width=1.0\columnwidth]{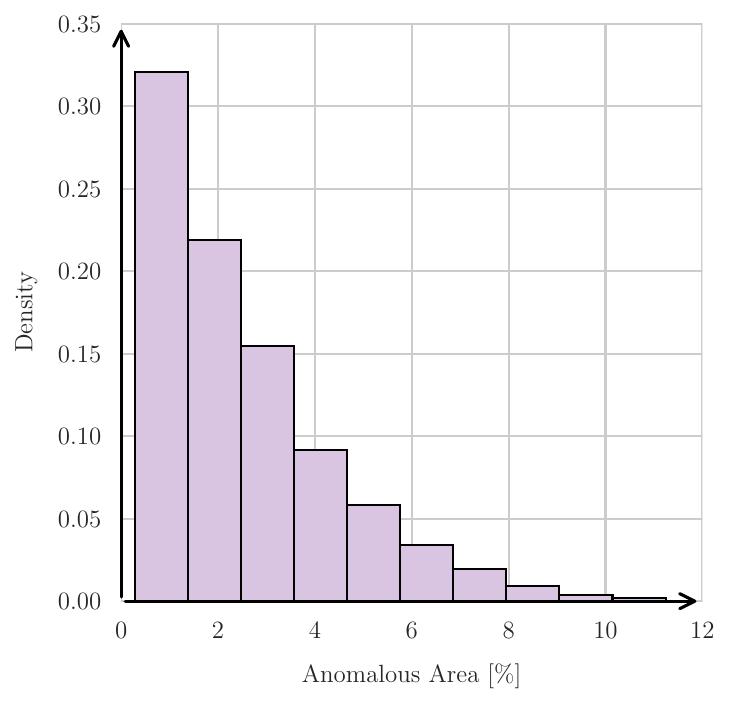}
    \caption{Anomalous Area Distribution in the generated synthetic dataset.
    }
    \label{fig:distr}
\end{figure}

Here, we provide details about the synthetic dataset generated for training our model. High-level statistics can be seen in Table~\ref{tab:syn_data}. The generated dataset contains all of the possible objects and backgrounds. Additionally, it contains 204 different anomalies, significantly more than current datasets (e.g. MVTec AD~\cite{bergmann2019mvtec} contains 73 different anomaly types). The generated anomalies are, in general, relatively small (on average, they account for 2.52\% of the image). In contrast, in MVTec AD, they occupy 4.39\% of the image. A more detailed visualisation of the anomaly area distribution is depicted in Figure~\ref{fig:distr}.

\section{Additional Ablation Studies}
\label{a:abl}

In this section, we present additional experiments that verify the design choices in AnomalyVFM. Most of the results are presented in Table~\ref{tab:abl_extra}

\begin{figure*}
    \centering
    \includegraphics[width=0.32\textwidth]{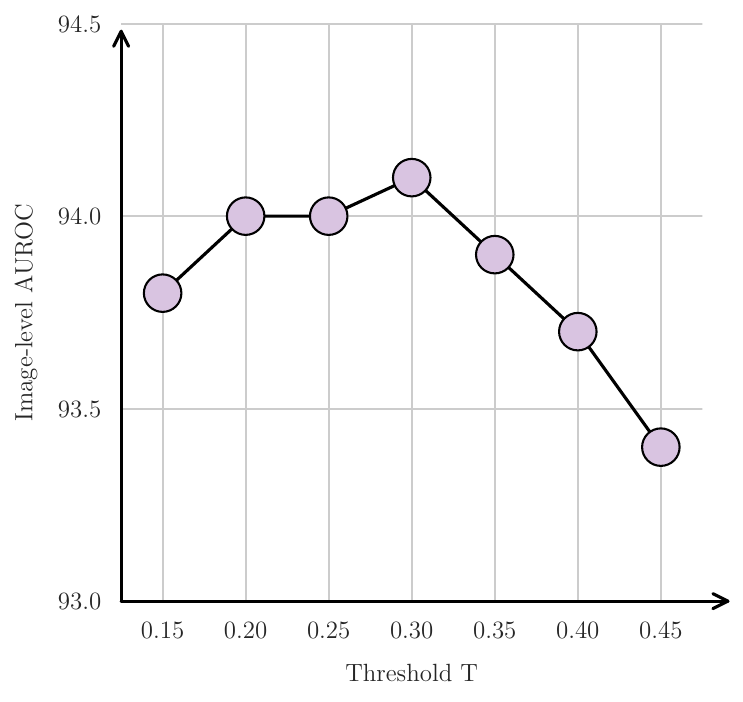}
    \includegraphics[width=0.32\textwidth]{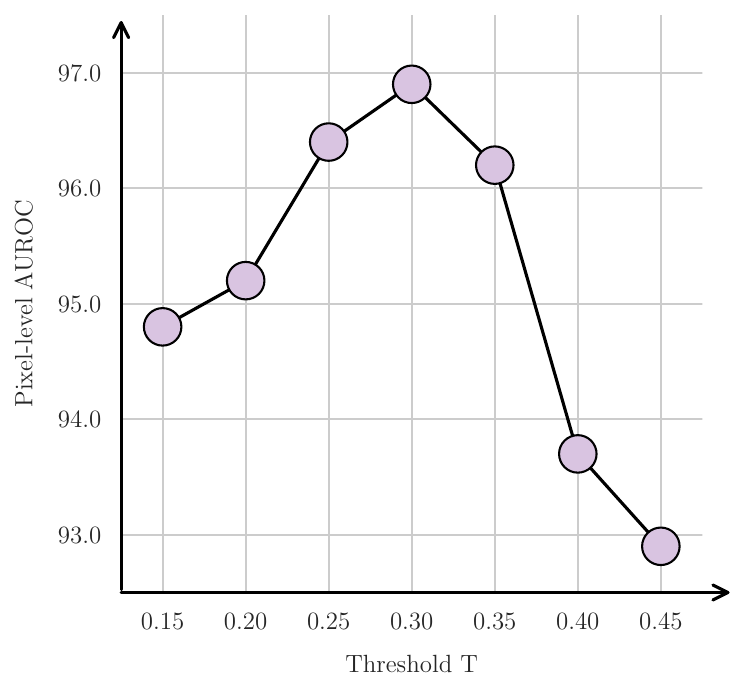}
    \includegraphics[width=0.32\textwidth]{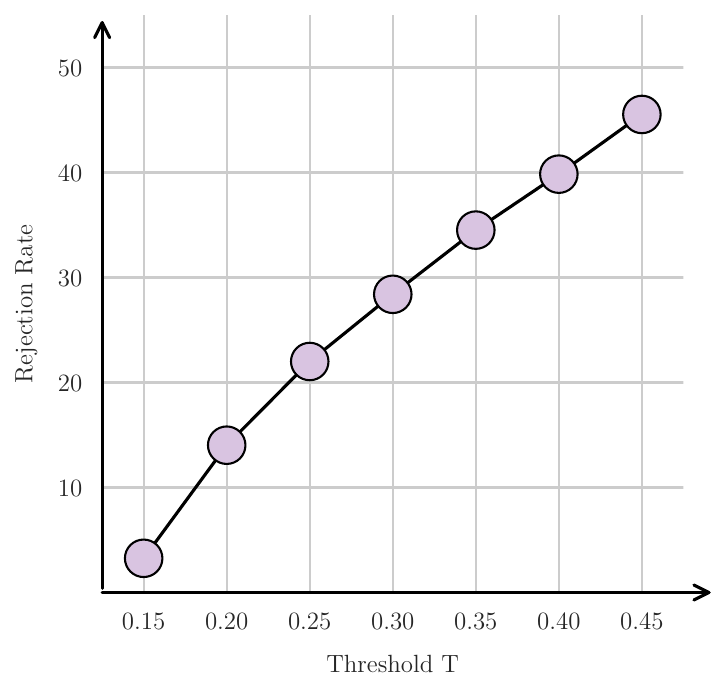}
    \caption{Model performance and rejection rate in relation to filtering threshold $T$.
    }
    \label{fig:filt_exp}
\end{figure*}

\noindent \textbf{Filtering Threshold} To verify the impact of the threshold $T$ used during dataset filtering, we re-filtered the dataset using various values of $T$. On top of the performance metrics, we also measured the rejection rate (i.e., the percentage of images discarded). The results and the rejection rate can be seen in Figure~\ref{fig:filt_exp}. The results show that the image-level AUROC is quite robust to the set threshold, while the pixel-level AUROC is more reliant on a correct choice of a threshold. At the default setting, the rejection rate is approximately $30\%$, showcasing that prompt adherence is far from a solved problem in generative models.

\begin{figure*}
    \centering
    \includegraphics[width=0.35\textwidth]{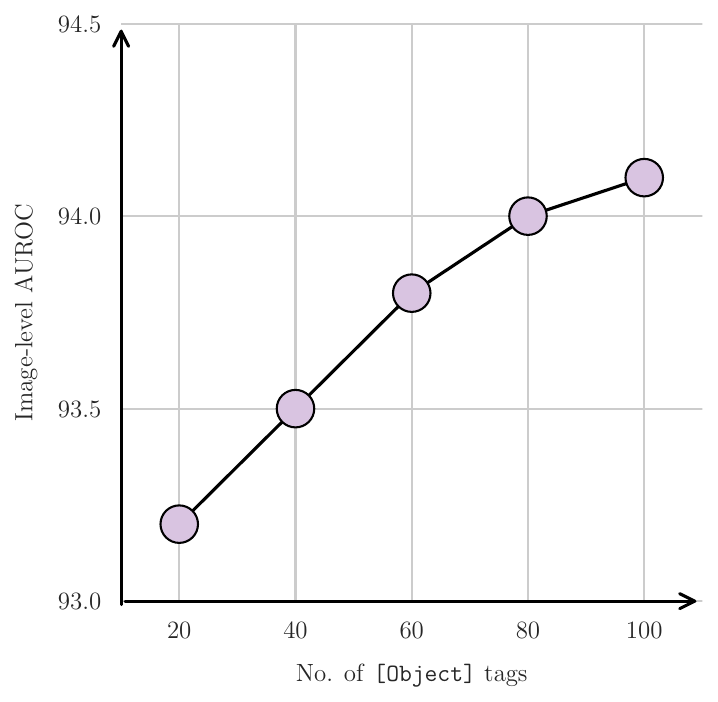}
    \includegraphics[width=0.35\textwidth]{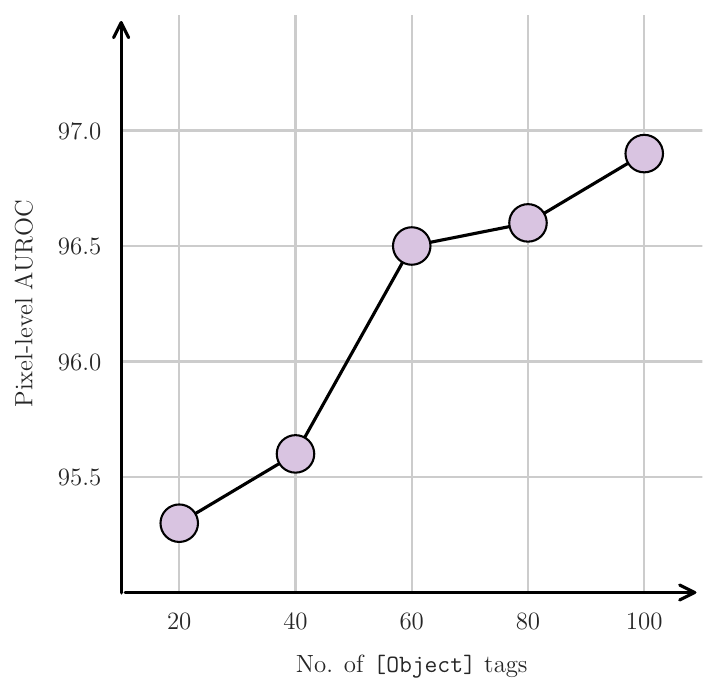}
    \caption{Model performance in comparison to the number of $\texttt{[Object]}$ tags.
    }
    \label{fig:obj_num}
\end{figure*}

\noindent \textbf{Number of $\texttt{[Object]}$ tags} To verify the importance of having a diverse dataset, we varied the number of $\texttt{[Object]}$ tags during the synthetic data generation phase. The results can be seen in Figure~\ref{fig:obj_num}. The results consistently rise with the number of $\texttt{[Object]}$ tags. The performance with 20 $\texttt{[Object]}$ tags is similar to the performance when AnomalyVFM is trained on MVTec AD~\cite{bergmann2019mvtec}, which has 15 different objects inside the dataset. We have not gone above 100 tags, as that is the list we initially generated with an LLM. In the future, we will increase this to see if the performance can be improved even further.

\begin{figure*}
    \centering
    \includegraphics[width=0.35\textwidth]{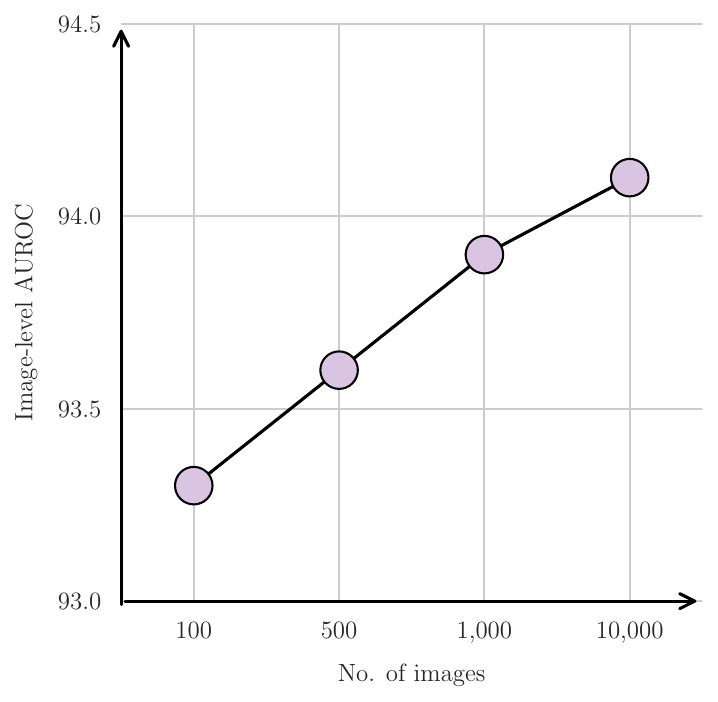}
    \includegraphics[width=0.35\textwidth]{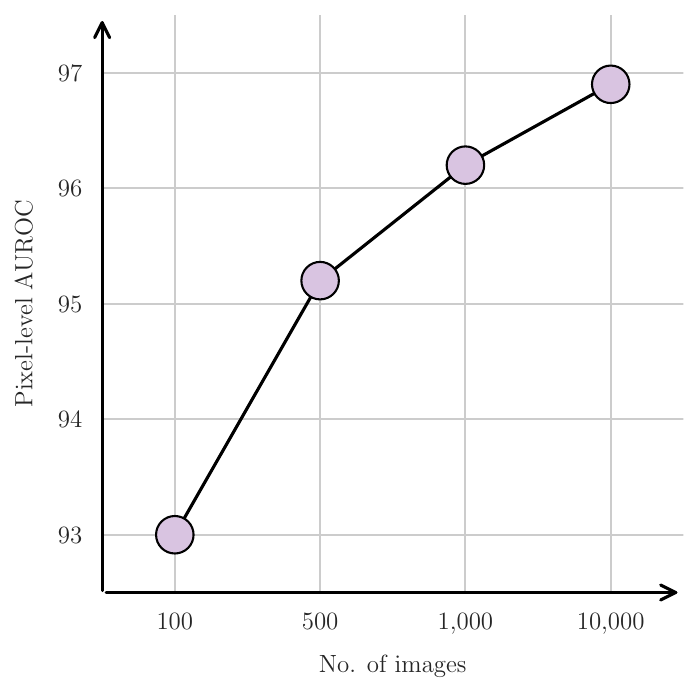}
    \caption{Model performance in comparison to the number of images in the training set.
    }
    \label{fig:img_num}
\end{figure*}

\noindent \textbf{Number of images} During all of our experiments, we used 10,000 generated images. To verify the importance of this, we have tried several different quantities: 100, 500, 1,000, and 10,000. The results are depicted in Figure~\ref{fig:img_num}. The performance increases steadily with each increment. We hypothesise that further scaling could improve performance even further. We have not done so to maintain a training set size similar to that of the related methods.

\noindent \textbf{LoRA Rank} To verify the robustness of the proposed method towards the rank of the LoRA adapters, we varied this parameter. More specifically, we decreased the rank to 32 and then increased it to 128.  Decreasing it leads to a decrease of $0.3$ p.\ p.\ in image-level AUROC and $0.3$ p.\ p.\ in pixel-level AUROC. Increasing the LoRA rank leads to no differences in image-level metrics, while the pixel-level AUROC decreases for $0.3$ p.\ p. This shows the robustness of the proposed method to this parameter.

\noindent \textbf{LoRA Positions} In the implementation, LoRA is added to query, value and projection layers inside the attention mechanism. This was done based on the insights from the open-source community on how to efficiently adapt image generation models. To verify the importance of this choice, we performed experiments with more layouts. All of the layouts keep a similar performance, showcasing robustness to this choice. The largest dip in performance is observed when LoRA adaptors are added to all linear layers. We assume this is the case as the model cannot pass the information globally but rather only locally.

\noindent \textbf{Model Size} RADIO has multiple model sizes. To verify the importance of this parameter, we exchanged it with a smaller (ViT-B) and larger (ViT-H) model. Using a smaller model leads to a decrease of $1.8$ p.\ p.\ in image-level AUROC and $1.2$ p.\ p.\ in pixel-level AUROC. A larger model leads to a decrease of $0.6$ p. p.\ in image-level AUROC and $1.2$ p.\ p.\ in pixel-level AUROC. This shows that ViT-L is the optimal choice. We also hypothesise that increasing the number of $\texttt{[Object]}$ tags and the total number of images would make ViT-H more optimal.

\section{Additional Qualitative Examples}
\label{a:qual}

\begin{figure*}[t]
    \centering
    \includegraphics[width=1.0\textwidth]{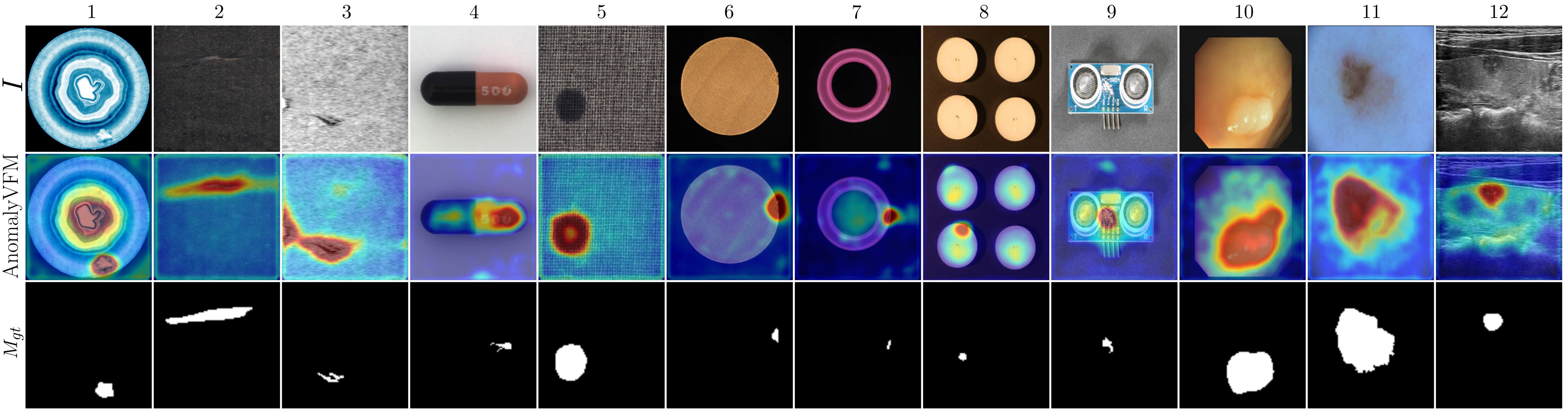}
    \caption{Qualitative examples of anomaly segmentation masks produced by AnomalyVFM. In the first row, the image is shown. In the next row, the anomaly segmentation produced by AnomalyVFM is depicted, and in the last row, the ground truth mask is depicted.
    }
    \label{fig:qual}
\end{figure*}

In this section, we add additional qualitative examples of anomaly segmentations produced by AnomalyVFM. The examples can be seen in Figure~\ref{fig:qual}. AnomalyVFM can detect anomalies across a wide range of objects.

\section{Image Generation Data}
\label{a:gen_data}

To enable reproducibility and to ensure transparency, we provide the list of $\texttt{[Object]}$, $\texttt{[Anomaly]}$ and $\texttt{[Texture]}$ used in the synthetic dataset generation. The lists of $\texttt{[Object]}$ and $\texttt{[Anomaly]}$ tags can be seen in Table~\ref{tab:obj_1} and Table~\ref{tab:obj_2}. The list of $\texttt{[Texture]}$ tags can be seen in Table~\ref{tab:bgs}.

\begin{table*}[h]
\centering
\caption{\texttt{[Object]} and \texttt{[Anomaly]} data used for synthetic data generation. Here are listed objects from A to L.}
\label{tab:obj_1}
\vspace{-8pt}
\setlength{\tabcolsep}{1mm}
\begin{tabular}{ll}
\toprule
\texttt{[Object]} & \texttt{[Anomaly]} \\ \midrule 
\textbf{\texttt{apple}} & [\texttt{bruised}, \texttt{wrinkled}, \texttt{rotten}, \texttt{moldy}, \texttt{dented}, \texttt{discolored}, \texttt{soft spots}] \\ 
\textbf{\texttt{apple slice}} & [\texttt{oxidized}, \texttt{bruised}, \texttt{dried out}] \\ 
\textbf{\texttt{asphalt}} & [\texttt{cracked}, \texttt{pitted}, \texttt{faded}, \texttt{eroded}, \texttt{oil-stained}, \texttt{uneven}] \\ 
\textbf{\texttt{ball}} & [\texttt{deflated}, \texttt{scuffed}, \texttt{punctured}, \texttt{faded}, \texttt{cracked surface}] \\ 
\textbf{\texttt{banana}} & [\texttt{bruised}, \texttt{overripe}, \texttt{blackened}, \texttt{split peel}, \texttt{mushy}, \texttt{spotted}] \\ 
\textbf{\texttt{battery}} & [\texttt{leaking}, \texttt{corroded}, \texttt{dented}, \texttt{faded label}] \\ 
\textbf{\texttt{belt}} & [\texttt{cracked leather}, \texttt{frayed edges}, \texttt{worn holes}, \texttt{peeling finish}] \\ 
\textbf{\texttt{bicycle}} & [\texttt{flat tire}, \texttt{rusty chain}, \texttt{scratched frame}, \texttt{worn seat}] \\ 
\textbf{\texttt{board game}} & [\texttt{torn box}, \texttt{missing pieces}, \texttt{faded board}, \texttt{bent cards}] \\ 
\textbf{\texttt{bread}} & [\texttt{stale}, \texttt{moldy}, \texttt{crumbling}, \texttt{burnt}, \texttt{hardened}, \texttt{soggy}] \\ 
\textbf{\texttt{brushed aluminum}} & [\texttt{scratched}, \texttt{dented}, \texttt{stained}, \texttt{faded}, \texttt{oxidized}, \texttt{pitted}] \\ 
\textbf{\texttt{butter}} & [\texttt{rancid}, \texttt{melted}, \texttt{discolored}, \texttt{greasy residue}, \texttt{hardened}] \\ 
\textbf{\texttt{car tire}} & [\texttt{bald tread}, \texttt{cracked rubber}, \texttt{punctured}, \texttt{worn sidewall}] \\ 
\textbf{\texttt{carbon fiber}} & [\texttt{frayed}, \texttt{chipped}, \texttt{cracked}, \texttt{delaminated}, \texttt{scratched}, \texttt{discolored}] \\ 
\textbf{\texttt{carrot}} & [\texttt{softened}, \texttt{cracked}, \texttt{dehydrated}, \texttt{spotted}, \texttt{moldy}, \texttt{bent}] \\ 
\textbf{\texttt{chair}} & [\texttt{scratched wood}, \texttt{stained cushion}, \texttt{wobbly leg}] \\ 
\textbf{\texttt{chalkboard}} & [\texttt{scratched}, \texttt{smudged}, \texttt{cracked}, \texttt{chipped}, \texttt{stained}, \texttt{uneven}] \\ 
\textbf{\texttt{cheese}} & [\texttt{moldy}, \texttt{dried out}, \texttt{cracked}, \texttt{discolored}, \texttt{sweating}, \texttt{crumbly}] \\ 
\textbf{\texttt{chocolate bar}} & [\texttt{melted}, \texttt{bloomed}, \texttt{crumbled}, \texttt{discolored}] \\ 
\textbf{\texttt{concrete}} & [\texttt{cracked}, \texttt{pitted}, \texttt{stained}, \texttt{eroded}, \texttt{chipped}, \texttt{weathered}] \\ 
\textbf{\texttt{cookies}} & [\texttt{crumbled}, \texttt{stale}, \texttt{burnt}, \texttt{moldy}] \\ 
\textbf{\texttt{cork}} & [\texttt{cracked}, \texttt{crumbled}, \texttt{stained}, \texttt{dried out}, \texttt{warped}, \texttt{pitted}] \\ 
\textbf{\texttt{corrugated metal}} & [\texttt{dented}, \texttt{rusted}, \texttt{bent}, \texttt{scratched}, \texttt{corroded}, \texttt{pitted}] \\ 
\textbf{\texttt{denim}} & [\texttt{frayed}, \texttt{torn}, \texttt{stained}, \texttt{faded}, \texttt{pilled}, \texttt{worn}] \\ 
\textbf{\texttt{doll}} & [\texttt{torn clothing}, \texttt{missing eye}, \texttt{stained}, \texttt{frayed hair}, \texttt{loose limbs}] \\ 
\textbf{\texttt{drill}} & [\texttt{worn chuck}, \texttt{scratched casing}, \texttt{broken switch}, \texttt{dented battery}] \\ 
\textbf{\texttt{egg}} & [\texttt{cracked}, \texttt{leaking}, \texttt{discolored shell}, \texttt{dented}, \texttt{rotten}, \texttt{thin shell}] \\ 
\textbf{\texttt{fabric}} & [\texttt{frayed}, \texttt{torn}, \texttt{stained}, \texttt{faded}, \texttt{pilled}, \texttt{snagged}] \\ 
\textbf{\texttt{fur}} & [\texttt{matted}, \texttt{shedding}, \texttt{stained}, \texttt{torn}, \texttt{faded}, \texttt{dull}] \\ 
\textbf{\texttt{garden hose}} & [\texttt{cracked}, \texttt{leaking}, \texttt{kinked}, \texttt{faded}] \\ 
\textbf{\texttt{garlic}} & [\texttt{sprouted}, \texttt{dried out}, \texttt{moldy}] \\ 
\textbf{\texttt{glasses}} & [\texttt{scratched lenses}, \texttt{bent frame}, \texttt{loose arms}, \texttt{cloudy lenses}] \\ 
\textbf{\texttt{gloves}} & [\texttt{frayed fingers}, \texttt{stretched out}, \texttt{stained}] \\ 
\textbf{\texttt{grape}} & [\texttt{wrinkled}, \texttt{moldy}, \texttt{shriveled}] \\ 
\textbf{\texttt{grill}} & [\texttt{rusty grates}, \texttt{blackened residue}, \texttt{scratched body}] \\ 
\textbf{\texttt{hammer}} & [\texttt{rusty}, \texttt{chipped}, \texttt{bent}, \texttt{dented}, \texttt{scratched}, \texttt{loose head}] \\ 
\textbf{\texttt{hat}} & [\texttt{faded color}, \texttt{stretched}, \texttt{frayed edges}] \\ 
\textbf{\texttt{headphones}} & [\texttt{frayed cable}, \texttt{scratched ear cups}, \texttt{loose padding}] \\ 
\textbf{\texttt{helmet}} & [\texttt{scratched shell}, \texttt{cracked foam}, \texttt{loose straps}] \\ 
\textbf{\texttt{hemp fabric}} & [\texttt{frayed}, \texttt{torn}, \texttt{faded}, \texttt{pilled}, \texttt{stained}, \texttt{snagged}] \\ 
\textbf{\texttt{jacket}} & [\texttt{broken zipper}, \texttt{faded color}, \texttt{torn lining}] \\ 
\textbf{\texttt{jeans}} & [\texttt{worn knees}, \texttt{frayed hem}, \texttt{ripped pocket}, \texttt{faded}] \\ 
\textbf{\texttt{key}} & [\texttt{bent}, \texttt{worn teeth}, \texttt{rusty}, \texttt{scratched surface}] \\ 
\textbf{\texttt{kite}} & [\texttt{torn fabric}, \texttt{bent frame}, \texttt{frayed string}, \texttt{missing tail}] \\ 
\textbf{\texttt{laminate}} & [\texttt{scratched}, \texttt{chipped}, \texttt{peeled}, \texttt{bubbled}, \texttt{stained}, \texttt{warped}] \\ 
\textbf{\texttt{lamp}} & [\texttt{flickering}, \texttt{scratched base}, \texttt{broken switch}] \\ 
\textbf{\texttt{laptop}} & [\texttt{scratched casing}, \texttt{cracked hinge}, \texttt{faded keyboard keys}] \\ 
\textbf{\texttt{lettuce}} & [\texttt{wilting}, \texttt{yellowing}, \texttt{rotting}] \\ 
\textbf{\texttt{light bulb}} & [\texttt{burnt out}, \texttt{cracked}, \texttt{blackened}, \texttt{loose filament}] \\ 
\textbf{\texttt{linen}} & [\texttt{wrinkled}, \texttt{stained}, \texttt{faded}, \texttt{torn}, \texttt{frayed}, \texttt{pilled}] \\
\bottomrule
\end{tabular}
\end{table*}

\begin{table*}[h]
\centering
\caption{\texttt{[Object]} and \texttt{[Anomaly]} data used for synthetic data generation. Here are listed objects from M to Z.}
\setlength{\tabcolsep}{1mm}
\label{tab:obj_2}
\vspace{-8pt}
\begin{tabular}{ll}
\toprule
\texttt{[Object]} & \texttt{[Anomaly]} \\ \midrule 
 
\textbf{\texttt{mesh}} & [\texttt{frayed}, \texttt{torn}, \texttt{snagged}, \texttt{discolored}, \texttt{brittle}, \texttt{stretched}] \\ 
\textbf{\texttt{milk carton}} & [\texttt{dented}, \texttt{leaking}, \texttt{stained}, \texttt{faded label}, \texttt{torn packaging}] \\ 
\textbf{\texttt{mirror}} & [\texttt{scratched}, \texttt{chipped edge}, \texttt{cloudy}, \texttt{stained surface}] \\ 
\textbf{\texttt{onion}} & [\texttt{sprouted}, \texttt{dried layers}, \texttt{rotting}] \\ 
\textbf{\texttt{orange}} & [\texttt{dried skin}, \texttt{moldy}, \texttt{bruised}, \texttt{discolored}] \\ 
\textbf{\texttt{paintbrush}} & [\texttt{frayed bristles}, \texttt{stiffened bristles}, \texttt{dried paint}] \\ 
\textbf{\texttt{paper}} & [\texttt{torn}, \texttt{wrinkled}, \texttt{stained}, \texttt{yellowed}, \texttt{brittle}, \texttt{moldy}] \\ 
\textbf{\texttt{parquet flooring}} & [\texttt{scratched}, \texttt{warped}, \texttt{faded}, \texttt{chipped}, \texttt{stained}, \texttt{dull}] \\ 
\textbf{\texttt{phone}} & [\texttt{cracked screen}, \texttt{scratched back}, \texttt{worn buttons}] \\ 
\textbf{\texttt{plastic}} & [\texttt{scratched}, \texttt{cracked}, \texttt{discolored}, \texttt{warped}, \texttt{brittle}, \texttt{faded}] \\ 
\textbf{\texttt{pliers}} & [\texttt{rusty}, \texttt{loose grip}, \texttt{scratched}, \texttt{chipped}, \texttt{stiff joint}] \\ 
\textbf{\texttt{plywood}} & [\texttt{warped}, \texttt{splintered}, \texttt{chipped}, \texttt{stained}, \texttt{delaminated}, \texttt{cracked}] \\ 
\textbf{\texttt{potato}} & [\texttt{sprouted}, \texttt{rotting}, \texttt{green spots}, \texttt{wrinkled}, \texttt{moldy}, \texttt{soft spots}] \\ 
\textbf{\texttt{rake}} & [\texttt{bent tines}, \texttt{rusty}, \texttt{loose handle}] \\ 
\textbf{\texttt{rattan}} & [\texttt{splintered}, \texttt{frayed}, \texttt{cracked}, \texttt{stained}, \texttt{brittle}, \texttt{discolored}] \\ 
\textbf{\texttt{rubber floor}} & [\texttt{cracked}, \texttt{brittle}, \texttt{discolored}, \texttt{stiffened}, \texttt{melted}, \texttt{torn}] \\ 
\textbf{\texttt{saw}} & [\texttt{rusty blade}, \texttt{dull teeth}, \texttt{chipped handle}, \texttt{bent blade}] \\ 
\textbf{\texttt{scarf}} & [\texttt{pilled fabric}, \texttt{snagged threads}, \texttt{stained}] \\ 
\textbf{\texttt{screwdriver}} & [\texttt{worn tip}, \texttt{rusty}, \texttt{scratched}, \texttt{bent}, \texttt{cracked handle}] \\ 
\textbf{\texttt{shoes}} & [\texttt{worn sole}, \texttt{scuffed leather}, \texttt{torn fabric}, \texttt{faded color}] \\ 
\textbf{\texttt{shovel}} & [\texttt{rusty blade}, \texttt{dented handle}, \texttt{worn grip}] \\ 
\textbf{\texttt{smooth ceramic tile}} & [\texttt{chipped}, \texttt{cracked}, \texttt{stained}, \texttt{crazed}, \texttt{dull}, \texttt{scratched}] \\ 
\textbf{\texttt{smooth glass}} & [\texttt{scratched}, \texttt{cracked}, \texttt{chipped}, \texttt{foggy}, \texttt{stained}, \texttt{shattered}] \\ 
\textbf{\texttt{smooth metal}} & [\texttt{rusted}, \texttt{scratched}, \texttt{dented}, \texttt{corroded}, \texttt{pitted}, \texttt{tarnished}] \\ 
\textbf{\texttt{smooth wood plank}} & [\texttt{cracked}, \texttt{splintered}, \texttt{warped}, \texttt{knotted}, \texttt{rotten}, \texttt{scratched}, \texttt{stained}] \\ 
\textbf{\texttt{socks}} & [\texttt{hole in toe}, \texttt{stretched elastic}, \texttt{faded}] \\ 
\textbf{\texttt{stainless steel}} & [\texttt{scratched}, \texttt{dented}, \texttt{stained}, \texttt{scuffed}, \texttt{fingerprinted}, \texttt{corroded}] \\ 
\textbf{\texttt{stone tile}} & [\texttt{chipped}, \texttt{cracked}, \texttt{eroded}, \texttt{stained}, \texttt{pitted}, \texttt{weathered}] \\ 
\textbf{\texttt{strawberry}} & [\texttt{moldy}, \texttt{bruised}, \texttt{shrinking}] \\ 
\textbf{\texttt{synthetic fiber}} & [\texttt{frayed}, \texttt{torn}, \texttt{stained}, \texttt{faded}, \texttt{pilled}, \texttt{melted}] \\ 
\textbf{\texttt{table}} & [\texttt{scratched surface}, \texttt{dented corner}, \texttt{water stains}] \\ 
\textbf{\texttt{tape measure}} & [\texttt{cracked casing}, \texttt{faded markings}, \texttt{stuck mechanism}] \\ 
\textbf{\texttt{teddy bear}} & [\texttt{ripped seam}, \texttt{matted fur}, \texttt{faded color}, \texttt{stained}, \texttt{missing stuffing}] \\ 
\textbf{\texttt{tent}} & [\texttt{torn fabric}, \texttt{bent poles}, \texttt{moldy spots}] \\ 
\textbf{\texttt{tomato}} & [\texttt{soft spots}, \texttt{cracked skin}, \texttt{moldy}] \\ 
\textbf{\texttt{toy car}} & [\texttt{scratched paint}, \texttt{missing wheel}, \texttt{cracked body}, \texttt{loose parts}] \\ 
\textbf{\texttt{TV remote}} & [\texttt{worn-out buttons}, \texttt{cracked case}, \texttt{faded labels}] \\ 
\textbf{\texttt{velvet}} & [\texttt{crushed}, \texttt{faded}, \texttt{stained}, \texttt{pilled}, \texttt{torn}, \texttt{frayed}] \\ 
\textbf{\texttt{wallet}} & [\texttt{worn edges}, \texttt{cracked leather}, \texttt{faded color}, \texttt{frayed stitching}] \\ 
\textbf{\texttt{wallpaper}} & [\texttt{peeled}, \texttt{torn}, \texttt{stained}, \texttt{faded}, \texttt{bubbled}, \texttt{wrinkled}] \\ 
\textbf{\texttt{watch}} & [\texttt{scratched face}, \texttt{broken strap}, \texttt{faded markings}, \texttt{cracked casing}] \\ 
\textbf{\texttt{whiteboard}} & [\texttt{scratched}, \texttt{stained}, \texttt{ghosting}, \texttt{cracked}, \texttt{faded}, \texttt{dented}] \\ 
\textbf{\texttt{window}} & [\texttt{scratched glass}, \texttt{cracked}, \texttt{foggy}] \\ 
\textbf{\texttt{woven mat}} & [\texttt{frayed}, \texttt{torn}, \texttt{faded}, \texttt{loose fibers}, \texttt{stained}, \texttt{worn}] \\ 
\textbf{\texttt{wrench}} & [\texttt{rusty}, \texttt{scratched}, \texttt{dented}, \texttt{worn edges}, \texttt{corroded}] \\ 
\textbf{\texttt{yo-yo}} & [\texttt{scratched}, \texttt{cracked}, \texttt{tangled string}, \texttt{chipped edge}] \\ 
\bottomrule
\end{tabular}
\end{table*}

\begin{table*}[h]
\centering
\setlength{\tabcolsep}{1mm}
\caption{\texttt{[Texture]} data used for synthetic dataset generation.}
\label{tab:bgs}
\vspace{-8pt}
\begin{tabular}{c}
\toprule
\texttt{[Texture]} \\ \midrule \textbf{\texttt{asphalt}}, \textbf{\texttt{bamboo}}, \textbf{\texttt{brick}}, \textbf{\texttt{brushed aluminum}}, \textbf{\texttt{canvas}} \\ \textbf{\texttt{carbon fiber}}, \textbf{\texttt{ceramic}}, \textbf{\texttt{chalkboard}}, \textbf{\texttt{clouds}}, \textbf{\texttt{concrete}} \\ \textbf{\texttt{cork}}, \textbf{\texttt{corrugated metal}}, \textbf{\texttt{denim}}, \textbf{\texttt{fabric}}, \textbf{\texttt{fleece}} \\ \textbf{\texttt{foam}}, \textbf{\texttt{fur}}, \textbf{\texttt{glass}}, \textbf{\texttt{granite}}, \textbf{\texttt{grass}} \\ \textbf{\texttt{gravel}}, \textbf{\texttt{hemp fabric}}, \textbf{\texttt{ice}}, \textbf{\texttt{laminate}}, \textbf{\texttt{linen}} \\ \textbf{\texttt{marble}}, \textbf{\texttt{mesh}}, \textbf{\texttt{metal}}, \textbf{\texttt{mirror}}, \textbf{\texttt{painted wall}} \\ \textbf{\texttt{paper}}, \textbf{\texttt{parquet flooring}}, \textbf{\texttt{pebbles}}, \textbf{\texttt{plastic}}, \textbf{\texttt{plywood}} \\ \textbf{\texttt{rattan}}, \textbf{\texttt{rubber}}, \textbf{\texttt{sand}}, \textbf{\texttt{snow}}, \textbf{\texttt{stainless steel}} \\ \textbf{\texttt{stone}}, \textbf{\texttt{synthetic fiber}}, \textbf{\texttt{tarpaulin}}, \textbf{\texttt{terrazzo}}, \textbf{\texttt{tile}} \\ \textbf{\texttt{velvet}}, \textbf{\texttt{wallpaper}}, \textbf{\texttt{whiteboard}}, \textbf{\texttt{wire mesh}}, \textbf{\texttt{woven mat}} \\\bottomrule
\end{tabular}
\end{table*}

\end{document}